\documentclass{article}
\usepackage{ijcai24}
\usepackage{times}
\usepackage{soul}
\usepackage{url}
\usepackage[hidelinks]{hyperref}
\usepackage[utf8]{inputenc}
\usepackage[small]{caption}
\usepackage{graphicx}
\usepackage{amsmath}
\usepackage{amsthm}
\usepackage{booktabs}
\usepackage[switch]{lineno}

\usepackage[ruled,vlined,linesnumbered]{algorithm2e}
\SetKw{KwTo}{to}
\SetKw{KwBy}{by}
\usepackage{url}
\usepackage{subfigure}

\usepackage[dvipsnames]{xcolor}

\definecolor{aureolin}{rgb}{0.99, 0.93, 0.0}
\usepackage{multirow}
\definecolor{lightgreen}{RGB}{217,255,179}

\newcommand{\drift}{\textsc{MEDVOC}}
\newcommand{\base}{\textsc{PLM}}

\newcommand{\best}[1]{\colorbox{lightgreen}{#1}}
\newcommand{\secondbest}[1]{\colorbox{aureolin}{#1}}

\usepackage{amssymb}%
\usepackage{pifont}%
\newcommand{\cmark}{\color{green}{\ding{51}}}%
\newcommand{\xmark}{\color{red}{\ding{55}}}%

\title{\drift{}: Vocabulary Adaptation for Fine-tuning Pre-trained Language Models on Medical Text Summarization}

\author{
Gunjan Balde\textsuperscript{\textsection}$^1$
\and
Soumyadeep Roy\textsuperscript{\textsection}$^{1,2}$\and
Mainack Mondal$^1$\And
Niloy Ganguly$^1$\\
\affiliations
$^1$Indian Institute of Technology Kharagpur\\
$^2$L3S Research Center, Germany
\emails
balde.gunjan0812@kgpian.iitkgp.ac.in, soumyadeep.roy9@iitkgp.ac.in,\{mainack, niloy\}@cse.iitkgp.ac.in
}

\begin{document}
\maketitle
\begingroup\renewcommand\thefootnote{\textsection}
\footnotetext{Equal contribution.}
\begingroup\renewcommand\thefootnote{}
\footnotetext{This is the author’s version of the work. It is posted here for your personal use. Not for redistribution. The definitive Version of Record was published in {\it Proceedings of the Thirty-Third International Joint Conference on Artificial Intelligence, (IJCAI-24)}, \url{https://doi.org/10.24963/ijcai.2024/683}}
\endgroup
\begin{abstract}
This work presents a dynamic vocabulary adaptation strategy, \drift, for fine-tuning pre-trained language models (PLMs) like BertSumAbs, BART, and PEGASUS for improved medical text summarization. In contrast to existing domain adaptation approaches in summarization, \drift{} treats vocabulary as an \textit{optimizable parameter}  and optimizes the PLM vocabulary based on \textit{fragment score}  conditioned only on the downstream task's reference summaries. Unlike previous works on vocabulary adaptation (limited only to classification tasks), optimizing vocabulary based on summarization tasks requires an extremely costly intermediate fine-tuning step on large summarization datasets. To that end, our novel \textit{fragment score}-based hyperparameter search very significantly reduces this fine-tuning time---from $450$ days to less than $2$ days on average. Furthermore, while previous works on vocabulary adaptation are often primarily tied to single PLMs, \drift{} is designed to be deployable across multiple PLMs (with varying model vocabulary sizes, pre-training objectives, and model sizes) ---bridging the limited vocabulary overlap between the biomedical literature domain and PLMs. \drift{} outperforms baselines by $15.74\%$ in terms of Rouge-L in zero-shot setting and shows gains of $17.29\%$ in high Out-Of-Vocabulary (OOV) concentrations.  Our human evaluation shows \drift{} generates \textit{more faithful} medical summaries ($88\%$ compared to $59\%$ in baselines).
\end{abstract}

\section{Introduction}

\begin{figure}[!ht]
    \centering
    \includegraphics[scale=0.35]{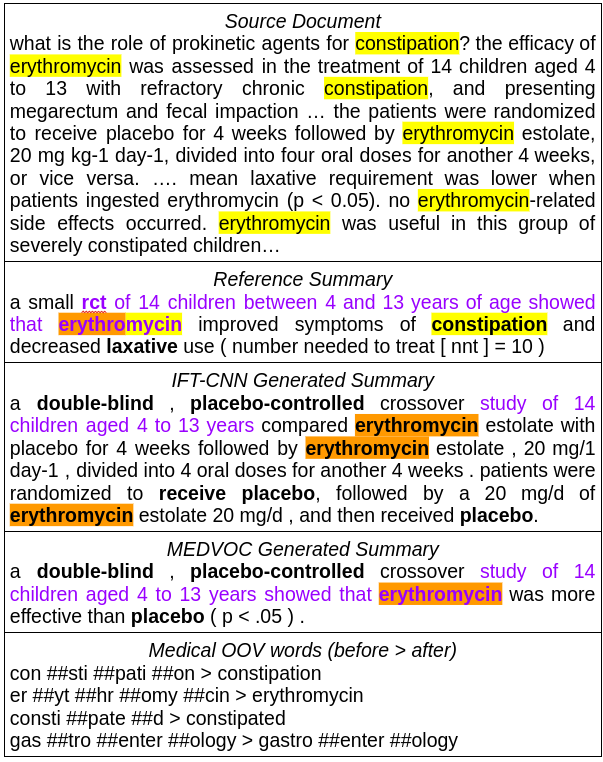}
    \caption{Illustrative example of BertSumAbs model from EBM dataset. Purple text color: indicates semantic or surface form overlaps with RS, Bold text: indicates medical (UMLS) concept-bearing words, Yellow highlight: OOV words that are ultimately added to the updated vocabulary, and Orange highlight: medical concept-bearing word(s) that overlap with reference summary. }%
    \label{fig:oov-summary-example}
\end{figure}

Medical text summarization is useful for many real-life use-cases %
such as summary generation of clinical records~\cite{kunwal-rizzo-et-al}, health-related queries~\cite{he-etal-2021-damo}, and radiology reports~\cite{dai-etal-2021-bdkg}. Most medical summarization approaches are based on pre-trained language models (PLMs) that are trained on text from open-domain sources. Thus, their performance is sub-optimal because they did not incorporate medical knowledge into their models~\cite{he-etal-2021-damo,zhang-etal-2021-leveraging-pretrained}. Domain adaptation approaches for summarization in general~\cite{fabbri-etal-2021-improving,laskar2022,xu-etal-2023-retrieval} and medical summarization~\cite{he-etal-2021-damo,healthinf22,zhu2023parameterefficient,xie2023survey}, in particular, have garnered reasonable research interest. %
 However, we identified two prominent research gaps in existing works. 
 
 First, despite poor domain similarity of $33\%$ in Figure~\ref{fig:oov_frequent}(a) between CNN/DailyMail~\cite{DBLP:conf/acl/SeeLM17} (open domain) and PubMed Abstracts Collection (medical domain), none of the domain adaptation approaches for medical summarization update the PLM's vocabulary during fine-tuning. 
 Figure~\ref{fig:oov-summary-example} demonstrates the challenges that arise when generating medical summaries without vocabulary adaptation. Figure~\ref{fig:oov_frequent}(b) demonstrates that medical concept-bearing words mostly lose their meaning due to poor representation because they are tokenized into four or more subwords. However, the undesirable tokenization actually happens at the decoder level during the generation of such medical concept-bearing words in a summary. Vocabulary adaptation is successful in the classification settings where it updates the PLM vocabulary adding a target domain-specific vocabulary~\cite{hong-etal-2021-avocado,xu-etal-2021-vocabulary,healthinf22}. Unfortunately, the vocabulary set construction algorithms of the classification setting are quite restrictive as they rely on fixed empirical thresholds~\cite{tai-etal-2020-exbert,hong-etal-2021-avocado} and fail to adapt themselves to a new, significantly different PLM architecture. %
  In this paper, we are \textbf{first to explore vocabulary adaptation techniques for summarization} and design the \drift{} fine-tuning strategy. However, adapting vocabulary adaptation strategies (earlier classification) to summarization (a generative setting) is non-trivial for \drift{}: (i) the decoder needs to be additionally trained, (ii) rare sub-words get included in the target domain-specific vocabulary, and (iii) absence of large-scale medical summarization datasets for intermediate fine-tuning purposes. We address these challenges in this work.

The second research gap is that most of the vocabulary adaptation till now is evaluated on a single PLM, as shown in Table~\ref{tab:related-work-compare}. To this end, as part of \drift{}, we develop an \textbf{efficient, dynamic vocabulary construction step} that adapts to any encoder-decoder-based (PLM) summarization model and target (downstream task) datasets. %
We treat vocabulary construction as a hyperparameter tuning step and show that optimizing the fragment score of reference summaries in a given setting closely resembles the same optimization as the downstream task performance. This helps to avoid the extremely time-consuming step of intermediate fine-tuning. Our fragment score-based hyperparameter search very significantly reduces this fine-tuning time, from $450$ days to $\sim45$ hours, averaged across three PLMs over four downstream medical summarization tasks. %

To address the two key research gaps, we need to first train the PLMs on the downstream medical summarization tasks. However, the size of the training datasets of the downstream medical summarization task is quite small, in the range of $700$ to $1525$ data points. Directly fine-tuning pre-trained models on such small \textit{target datasets} could lead to sub-optimal performance~\cite{phang2019sentence}. The standard approach relies on introducing an intermediate fine-tuning stage using large datasets~\cite{chang-lu-2021-rethinking-intermediate,suresh-etal-2023-intermediate}. In this work, we show that the \textbf{task of biomedical paper title generation serves as a good intermediate fine-tuning task}~\cite{fabbri-etal-2021-improving} for medical text summarization.%

\drift{} outperforms baselines by $15.74\%$, $4.80\%$, and $5.99\%$ in zero-shot, few-shot, and full dataset settings respectively on average across four medical summarization datasets and three PLMs such as BertSumAbs~\cite{liulapata2019text}, BART~\cite{lewis2020bart} and PEGASUS~\cite{pmlr-v119-zhang20ae} in terms of Rouge-L. \drift{} produces more informative ($5.99\%$ Rouge-L improvement on average) and more faithful medical summaries ($5.96\%$ Concept Score improvement on average). We observe gains of  $10.81$\% and $17.29$\% across challenging scenarios of long-form medical summary generation and reference summaries with high out-of-vocabulary (OOV) concentration respectively. %
\section{Related Works}\label{sec:sec_RW}
To the best of our knowledge, this is the first work to explore vocabulary adaptation strategies for the summarization task. However, recent works have explored vocabulary adaption strategies for classification tasks and domain adaptation techniques for summarization, which we present in Table~\ref{tab:related-work-compare}.  

\noindent \textbf{Vocabulary Adaptation Strategies for Classification.} To handle the vocabulary mismatch issue, BioBERT~\cite{cite:BioBert} and Paul\textit{ et al.,}~\shortcite{spaul_legal}, retrained the model from scratch using a domain-specific corpus and showed performance improvement over the base pre-trained models. While other works like VOLT~\cite{xu-etal-2021-vocabulary} and AVocaDo~\cite{hong-etal-2021-avocado} aim to optimize the model's vocabulary by adding a set of subwords to the existing vocabulary using some utility scoring function. AVocaDo uses a fragment score-based~\cite{rust-etal-2021-good} threshold, whereas T-DNA~\cite{diao-etal-2021-taming} selects n-grams with high Pointwise Mutual Information and iteratively merges them. exBERT~\cite{tai-etal-2020-exbert} adopts an ad-hoc approach to determine the size of $V_{TGT}$ and fix it as $17K$ ($56$\% of pre-trained vocabulary size), and does not perform any vocabulary optimization.
However, all these vocabulary adaptation works have two major drawbacks: (i) they are limited to classification tasks, and (ii) they show results on a single model type and their algorithm is not flexible enough to handle different model types. To the best of our knowledge, this is the first work that explores vocabulary adaptation strategies for summarization over multiple PLMs. We focus on encoder-decoder-based PLMs because it is computationally infeasible to re-train medical LLMs~\cite{chen2023meditron,wu2023pmcllama}.

\begin{table}[t]
    \centering
    \scalebox{0.7}{
    \begin{tabular}{p{3.5cm}p{3.25cm}p{2.5cm}cc}
    \hline
    \textbf{Related Works}  & \textbf{PLMs} & \textbf{Task} & \textbf{VA} & \textbf{IFT} \\ \hline
    Tai \textit{et al.,}~\shortcite{tai-etal-2020-exbert}  & BERT & Classification & \cmark{} & \xmark{}\\ 
    Dioa \textit{et al.,}~\shortcite{diao-etal-2021-taming}  & RoBERTa & Classification & \xmark{} & \xmark{} \\ 
    Hong \textit{et al.,}~\shortcite{hong-etal-2021-avocado}  & BERT & Classification & \cmark{} & \xmark{}  \\
    Lamproudis \textit{et al.,}~\shortcite{healthinf22}  & BERT & Classification & \cmark{} & \xmark{}\\ 
    Xu \textit{et al.,}~\shortcite{xu-etal-2023-retrieval} & BERT & Classification & \xmark{} & \xmark{} \\ \hline
    Xu \textit{et al.,}~\shortcite{xu-etal-2021-vocabulary} & Transformer-Big & Machine Translation & \cmark{} & \xmark{} \\
    Liu \textit{et al.,}~\shortcite{liu2023task} & GPT-2, BART & Question Answering & \cmark{} & \xmark{} \\
    Fabbri \textit{et al.,}~\shortcite{fabbri-etal-2021-improving}  & BART-L & Summarization & \xmark{} & \cmark{}   \\ 
    Xie \textit{et al.,}~\shortcite{xie2022pre} & BERT & Summarization & \xmark{} & \xmark{} \\  \hline
    \drift{} (Ours)  & BART-L, Pegasus-L, BertSumAbs & Summarization & \cmark{} & \cmark{}  \\ \hline
    
    \end{tabular}
    }
    \caption{Comparison of experimental setup with related works. \textit{VA} stands for \textit{Vocabulary Adaptation} and \textit{IFT} stands for \textit{Intermediate fine-Tuning}. `-L' refers to the \textit{Large} model variant.} %
    \label{tab:related-work-compare}
\end{table}
\section{Proposed Methodology} \label{sec:methods}
Here, we describe the \drift{} fine-tuning strategy for adapting PLMs to medical text summarization tasks in Figure~\ref{fig:medvoc-method}. We present the dynamic vocabulary construction step of \drift{} in Section~\ref{sec:vocab-adapt-method}, and then explain the intermediate fine-tuning details in Section~\ref{sec:ifts}. %

\begin{figure}[!hb]
    \centering
    \includegraphics[width=0.48\textwidth]{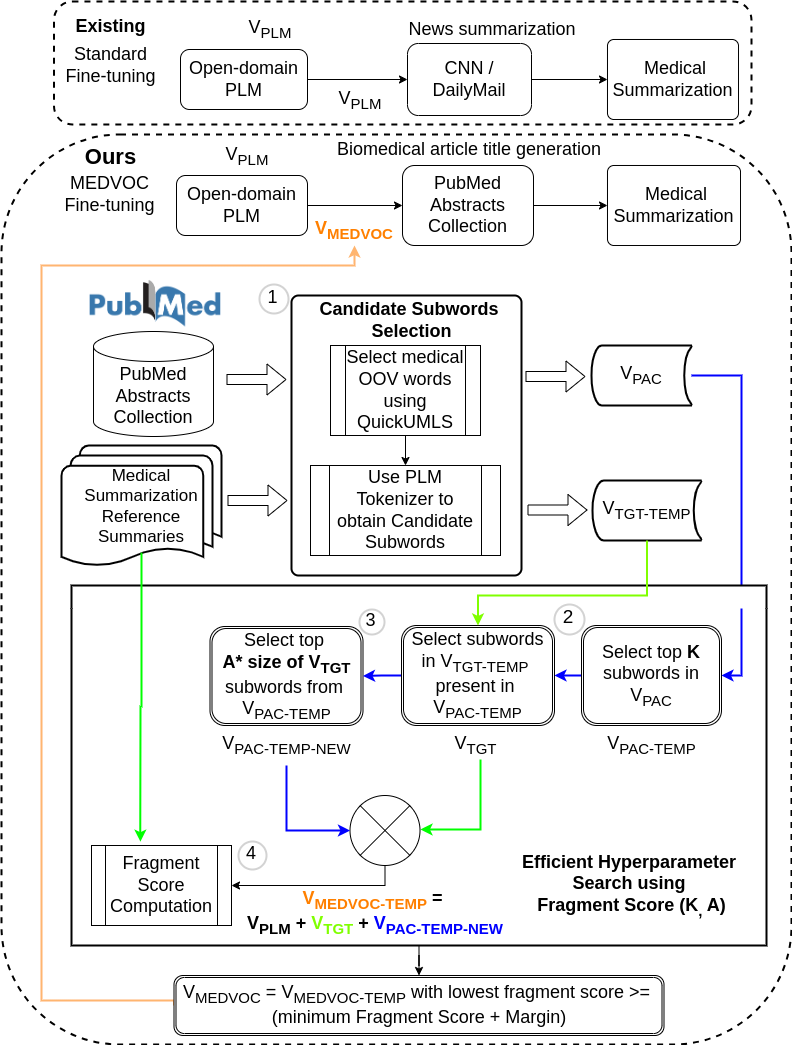}
    \caption{Methodological overview of \drift{} and existing fine-tuning strategy.}
    \label{fig:medvoc-method}
\end{figure}

\subsection{Dynamic Task-Aware Vocabulary Adaptation}\label{sec:vocab-adapt-method}
The key challenge is that the base PLM's vocabulary ($V_{\text{PLM}}$) remains unchanged during intermediate fine-tuning. Since this model vocabulary is obtained from training on open-domain data, we observe that ${V_{\text{PLM}}}$ misses (important) medical terms (Figure~\ref{fig:oov_frequent}(b)) present in the target dataset, which occurs quite frequently across datasets considered in this study. This causes the \base{} model tokenizer to split relevant medical terms into too many meaningless subwords, which ultimately is not able to capture the term semantics. This phenomenon is well-observed in prior studies w.r.t classification tasks~\cite{xu-etal-2021-vocabulary,hong-etal-2021-avocado} and results in poor downstream task performance. Therefore, we update $V_{\text{PLM}}$ by adding a set of target domain-specific subwords to the vocabulary, we refer to this updated model vocabulary as $V_\text{{\drift{}}}$. %

\subsubsection{Candidate Subwords Generation}
We first identify a set of words present in reference summaries (RS) of target tasks that are poorly tokenized, i.e., split into more than $3$ subwords. However, this resulted in poor coverage as it constituted a small fraction of $21.99\%$ out of the total set of out-of-vocabulary (OOV) words on average, i.e., words split into more than one subword. Therefore, we also include medically relevant OOV words. We use the \textit{matcher.match} function of QuickUMLS~\cite{soldaini2016quickumls} tool, with the parameters - (i) similarity measure as `$cosine$', and (ii) similarity threshold at $95\%$, to obtain the set of such medically relevant words. %
We run the PLM's tokenization scheme on these selected words for the target task to obtain $V_{\text{TGT-TEMP}}$. We apply the same procedure on the source documents of the Pubmed Abstracts Collection (PAC) dataset to obtain $V_{\text{PAC}}$ (see Section~\ref{sec:pac-construct} for more details). 

\subsubsection{Vocabulary Construction using Target Datasets (V\textsubscript{TGT})}
We observe that $V_{\text{TGT-TEMP}}$ contain subwords that are infrequent in PAC and it is well known that adding infrequent downstream task-specific subwords may lead to the rare word problem~\cite{Schick_Schütze_2020,hong-etal-2021-avocado}. Since PAC is used for intermediate fine-tuning, these infrequent words would appear in fewer contexts during training, and thus lead to a sub-optimal (noisier) representation. Therefore, we only consider subwords in $V_{\text{TGT-TEMP}}$ that overlap with the top $K$ subwords of $V_{\text{PAC}}$, thus mitigating the rare word issue. This value of $K$ is the first hyperparameter for \drift{}. However, we \textit{empirically observe that the size of $V_{\text{TGT}}$ is quite small as compared to the PLM's vocabulary size} ($3.66$\%, $1.58$\%, and $0.13$\% for BertSumAbs, BART, and PEGASUS respectively). This leads to marginal performance improvement as the added sub-words are overshadowed by the PLM vocabulary during summary generation. %

\subsubsection{Optimal Subset Selection from V\textsubscript{PAC}}
Since this step is upper-bounded by the size of $V_{\text{PAC}}$, which is quite large (3 times the model vocabulary), a large vocabulary causes parameter explosion and token sparsity problems, which hurts model learning~\cite{allison2006another,xu-etal-2021-vocabulary}. Therefore, we put an upper limit equal to the $|V_{\text{PLM}}|$. This also makes sure that the added vocabulary size does not exceed the PLM vocabulary size. Thus, determining the optimal subset size of $V_{\text{PAC}}$ lies in a tradeoff between model vocabulary size (large enough w.r.t PLM vocabulary) and model performance (small enough to not degrade downstream task performance). Therefore, we also include top $P$ subwords from $V_{\text{PAC}}$, where $P = A \times |V_{\text{TGT}}|$. The value of $A$ thus forms the second hyperparameter for \drift{}. 

\subsubsection{Efficient Hyperparameter Search using Fragment Score} 
The standard hyperparameter search step to find optimal values of $A$ and $K$ based on downstream task performance is extremely time-consuming. This is because it involves an additional intermediate fine-tuning step using PAC ($312K$ data points) that takes $45$ hours to run across $3$ Tesla V100 $32$ GB GPUs (averaged across the three PLMs over four datasets). An exhaustive grid search over $240$ settings takes $450$ days. Our efficient fragment score-based hyperparameter search performs the same in $2$ days on average, thus leading to a $240$x speedup. Upon extensive evaluation, we observe that by optimizing the fragment score for a given model type and dataset, the downstream summarization task performance also gets optimized.

Fragment score~\cite{hong-etal-2021-avocado,rust-etal-2021-good} is defined as the average number of subwords that a word gets tokenized on average by the base model tokenizer on the target dataset. Since computing the fragment score requires the downstream task's reference summaries and is independent of the intermediate fine-tuning step, the time taken for hyperparameter search drastically reduces to only a few hours. We find that the lowest time taken is $1 $hour and $30$ minutes on a single core of Intel i5 12-core CPU for BertSumAbs on the CHQSum dataset, whereas the highest time taken is $5$ hours and $45$ minutes for PEGASUS on the EBM dataset. Therefore, the speedup observed by \drift{} is proportional to the grid search space used for hyperparameter optimization. Our final vocabulary $V_{\text{\drift{}}}$ comprises subwords that lead to fragment scores within a certain range of best-achievable (minimum) fragment scores on the target dataset's reference summaries. Algorithm~\ref{algo:drift_creator} further explains the dynamic vocabulary construction step.

\newcommand\mycommfont[1]{\small\ttfamily\textcolor{blue}{#1}}
\SetKwInput{KwInput}{Input}
\SetKwInput{KwOutput}{Output}
\SetKwInput{KwData}{Initialization}
\SetCommentSty{mycommfont}
\begin{algorithm}[t]
\small
\caption{Vocabulary Construction of \drift{}}\label{algo:drift_creator}

\DontPrintSemicolon \;
    \KwInput{Pre-trained Vocabulary ($V_{\text{PLM}}$), Model tokenizer type $T$, Source documents of Pubmed Abstracts Collection $d_{\text{PAC}}$, Reference Summaries of Target dataset $d_{\text{TGT}}$}
    \KwOutput{\drift{} vocabulary --$V_{\text{MEDVOC}}$}
    \KwData{
        $K$: Selecting Top-K subwords in $V_{\text{PAC}}$, $A$: Factor over size of $V_{\text{TGT}}$, $Margin$: $0.04$
    }

    \SetKwFunction{FFragmentScore}{FragmentScore}
    \SetKwProg{Fn}{Function}{:}{}
    
    \Fn{\FFragmentScore{$d_{train}, V$}}{
     $f_{C}(V) \gets \frac{\text{total count of subwords tokenized by } V}{\text{word count in } d_{train}}$ \;
      
        \KwRet\  $f_{C}(V)$ \;
    }

    $V_{\text{TGT-TEMP}} \gets$ CandidateSubwordsSelection $(d_{TGT}, T)$ \;
    
    $V_{\text{PAC}} \gets$ CandidateSubwordsSelection $(d_{PAC}, T)$ \;

\For{$A \gets 0.25$ \KwTo $10$ \KwBy $0.25$}{
    \For{$K \gets 5000$ \KwTo min$(V_{\text{PLM}}, V_{\text{PAC}})$ \KwBy $5000$}{
        $V_{\text{PAC-TEMP}} \gets V_{\text{PAC}}[0:K]$ \tcp*[l]{Select top-K subwords}

        $V_{\text{TGT}} \gets V_{\text{TGT-TEMP}} \cap V_{\text{PAC-TEMP}}$ \tcp*[l]{Mitigating the rare word issue}

        $P =$ min$(V_{\text{PLM}}, A \cdot |V_{\text{TGT}}|)$ \tcp*[l]{Size to sample from $V_{\text{PAC-TEMP}}$}

        $V_{\text{PAC-TEMP-NEW}} \gets V_{\text{PAC}}[0:P]$ \tcp*[l]{Select top-$P$ subwords from $V_{\text{PAC}}$}

        $V_{\text{MEDVOC-TEMP}} \gets V_{\text{PLM}} \cup V_{\text{TGT}} \cup V_{\text{PAC-TEMP-NEW}}$

        $fragment\_score(A, K) \gets \text{\FFragmentScore}(d_{\text{TGT}}, V_{\text{MEDVOC-TEMP}})$ \;
    }
}
    $ \text{minFragScore} \gets$ Minimum fragment score across all values of $A, K$ \;
    
    $ V_{\text{MEDVOC}}  \gets V_{\text{MEDVOC-TEMP}}$  with smallest vocabulary size  with fragment score within a $Margin$ of $minFragScore$ \;

    \KwRet\  $ V_{\text{MEDVOC}}$ \;
    
\end{algorithm}

\noindent \textbf{Time Complexity of \drift{}.}  The candidate subword selection step as well as the fragment score computation is proportional to the size of the input corpus ($d_{\text{PAC}}$ and $d_{\text{TGT}}$). The hyperparameter search space is dependent on a constant set of values of $A$, while $K$ is conditioned on $|V_{\text{PAC}}|$, with the upper limit being $|V_{\text{PLM}}|$, which results in time complexity of  O($|V_{\text{PLM}}| * |d_{\text{TGT}}|$). The time complexity of \drift{} is O($|d_{\text{PAC}}|$) + O($|d_{\text{TGT}}|$) + O($|V_{\text{PLM}}| * |d_{\text{TGT}}|$). Since, our target task datasets have limited size ($|d_{\text{TGT}}|$ ranges between $700$ to $1525$ documents) and  $|d_{\text{TGT}}| << |d_{\text{PAC}}|$. The final time complexity is O($|V_{\text{PLM}}| * |d_{\text{TGT}}|$) +  O($|d_{\text{PAC}}|$). 

\subsection{Intermediate Fine-Tuning with Biomedical Article Title Generation} \label{sec:ifts}
Intermediate fine-tuning (IFT) is known to help PLMs when the downstream task has limited training (fine-tuning) data~\cite{chang-lu-2021-rethinking-intermediate,fabbri-etal-2021-improving}. In our case, the training dataset sizes range between $700$ and $1525$ data points (see Table~\ref{tab:dataset-stats}), which is too small for training purposes and would easily overfit the PLMs, leading to poor performance. However, large-scale summarization datasets (similar to CNN/DailyMail in the open domain) are required for meaningful intermediate fine-tuning, which is absent in the medical domain. We show that \textit{biomedical article title generation} satisfies the properties of a good intermediate fine-tuning task before fine-tuning on the downstream task of medical abstractive text summarization. %
Given that a good intermediate task aims to capture the knowledge or characteristics of the target task~\cite{chang-lu-2021-rethinking-intermediate,suresh-etal-2023-intermediate}, we use PAC as an intermediate fine-tuning task because it closely reflects key characteristics of the downstream summarization datasets (see Section~\ref{sec:ablation-analysis}).

\noindent \textbf{Fine-tuning Details.} Intermediate fine-tuning is performed using PAC (once for every PLM), whereas standard fine-tuning is done using downstream summarization tasks (once for every target dataset and PLM combination). Except for difference in the training dataset, we follow the standard fine-tuning procedure for the summarization task. We observe a marginal increase in the model's parameter count on adopting the \drift{} strategy as only the embedding matrix corresponding to the added vocabulary needs to be additionally trained. The parameter count increments by $0.15$\%, $1.15$\% and $1.59$\% in case of PEGASUS, BART, and BertSumAbs.
\section{Experimental Setup}\label{sec:expt-setup}
We describe datasets and evaluation metrics, followed by baselines and training details (more details in Appendix~\ref{appendix:expt-setup}).

\subsection{Target Task Datasets}\label{sec:target-task-dataset}
We evaluate \drift{} on two medical document summarization and two medical question summarization tasks. We provide detailed description of datasets in Appendix~\ref{appendix:data_cleaning}.

\noindent \textbf{Medical document summarization:} In \textit{BioASQ}~\cite{cite:BioASQ} and \textit{EBM}~\cite{DBLP:conf/acl-alta/MollaS11a}, each data point contains a query and PubMed (a biomedical database) abstract as the source document (SD) and an answer to the query as a reference summary (RS). 

\noindent \textbf{Medical question summarization:} \textit{MeQSum}~\cite{ben-abacha-demner-fushman-2019-summarization} and \textit{CHQSum}~\cite{yadav2022chq} contain consumer health questions posed by (non-medical) users as the SD and a short question (manually curated by medical experts) as the RS. \textit{MeQSum} and CHQSum comprise questions provided by the U.S National Library of Medicine and the Yahoo! Answers L6 corpus.

\begin{table}[!ht]
\centering
\scalebox{0.7}{
\setlength{\tabcolsep}{0.1cm}
\begin{tabular}{ccccccccc} 
\hline
\textbf{Dataset} &  \multicolumn{3}{c}{\textbf{Document count}} & \multicolumn{2}{c}{\textbf{Word count}} &\multicolumn{3}{c}{\textbf{OOV \%}} \\
{} & \textbf{Train} &\textbf{Val} & \textbf{Test} & \textbf{SD} & \textbf{RS} &  \textbf{BSA} &  \textbf{BART} & \textbf{PEGASUS} \\
\hline
CNN/Dailymail & 287,227 & 13,368 & 11,490 & 700 & 57 & 7.5 & 11.0 & 17.4  \\
PAC-Summ & 391,618 &  21,754 &  21,756 & 276  & 15  & 25.0 & 44.4 & 26.7\\ \hline
EBM & 1423 & 209 & 424  & 298  & 58  &  14.3 & 11.5 & 18.2\\
BioASQ &  1525 & 491 & 496 & 505  & 40  & 20.0 & 9.4 & 26.0\\ 
MeQSum &  700 & 150 & 150 & 70 & 12  & 12.5 & 5.7 & 16.7 \\ 
CHQSum & 1000 &  107& 400 & 184 & 12& 8.3 & 6.3 & 12.5\\
\hline
\end{tabular}}
\caption{Dataset statistics of intermediate fine-tuning datasets (CNN/DailyMail, PAC) and downstream medical summarization datasets. \textit{OOV\%} refers to the median fraction of unigrams in RS that are absent from the \base{} vocabulary.} %
\label{tab:dataset-stats}
\end{table}

\begin{figure}[!ht]
    \centering
    \subfigure[]{\includegraphics[width=0.25\textwidth]{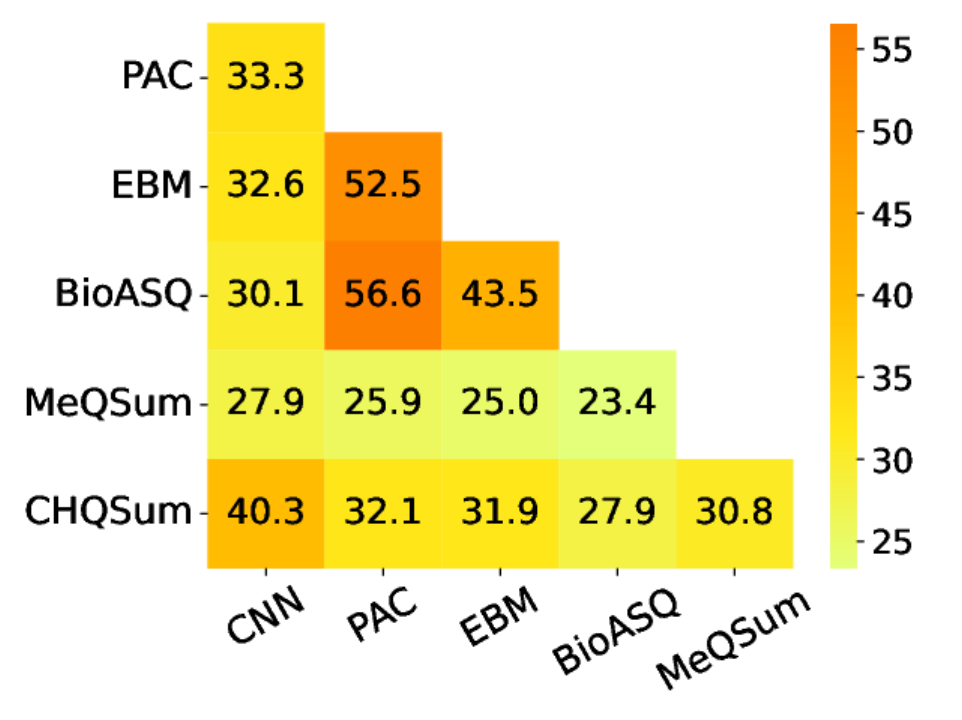}}
    \subfigure[]{\includegraphics[height=0.17\textwidth,width=0.17\textwidth]{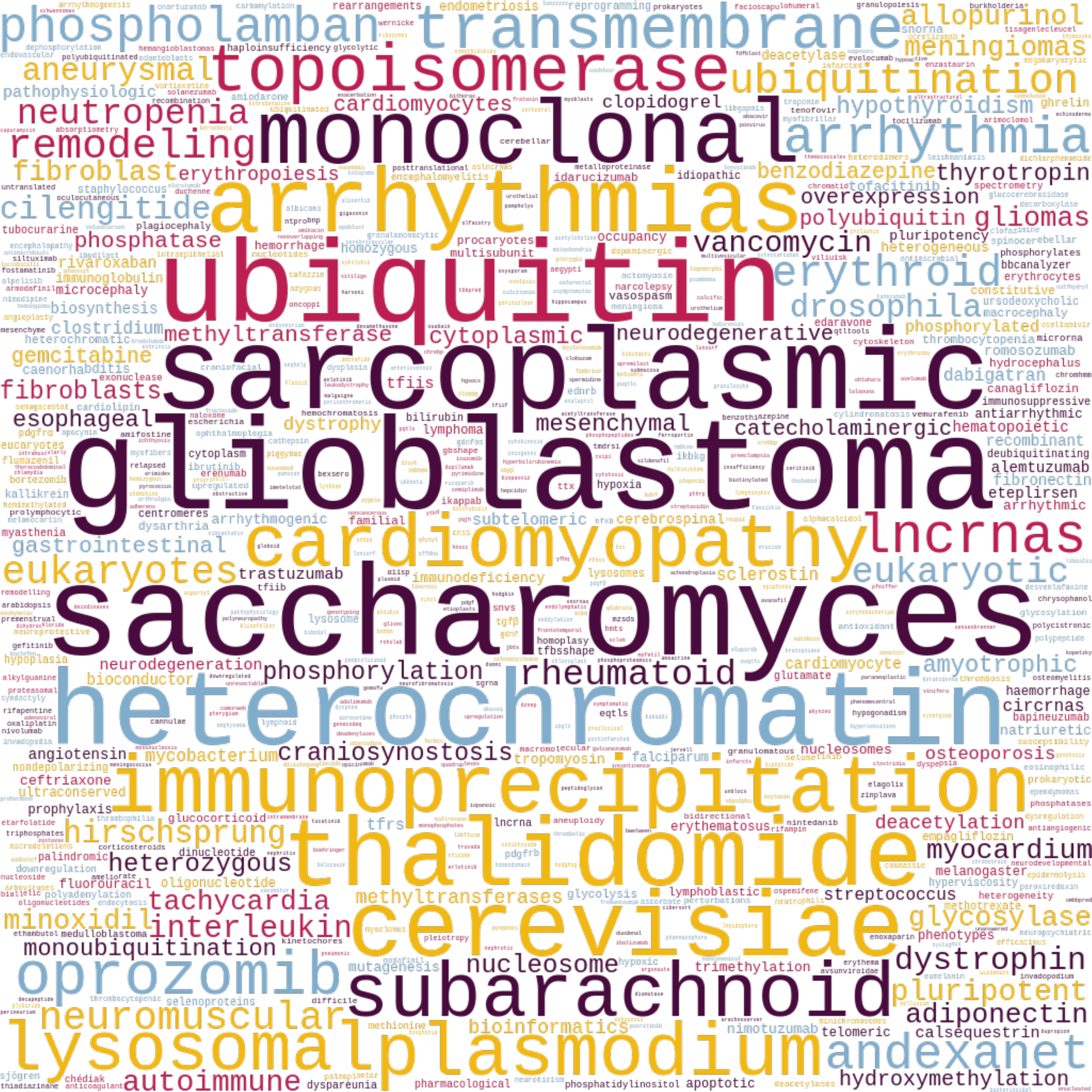}}
    \caption{(a) Heatmap to show vocabulary overlap among different training datasets, computed based on the overlap between the top 10K most frequent words in each dataset. CNN corresponds to CNN/DailyMail dataset. (b) Words in the BioASQ dataset across three PLMs are split into four or more subwords; we observe that most of them are medical terms.} %
    \label{fig:oov_frequent}
\end{figure} 

\subsection{Evaluation Metrics}\label{sec:eval-metrics}
We report Rouge~\cite{lin-2004-rouge} and BertScore~\cite{ZhangKWWA20} (BSr) to assess general summarization quality. We use Rouge-L as the main evaluation metric in line with prior works~\cite{yuan-etal-2022-biobart,zhang2023biomedgpt}. \textit{Concept F1-score} (CSr) is used to measure the faithfulness of medical summaries~\cite{zhang2023famesumm} and is computed as the overlap of UMLS medical concepts between the generated and reference summaries. We extract the medical concepts using the QuickUMLS~\cite{soldaini2016quickumls} tool. We also discuss an additional metric (MedRouge) and parmters of Rouge in Appendix~\ref{appendix:ROUGE}. %

\subsection{Baseline Models}\label{sec:baseline-models}
We provide further implementation details for the baseline models in Appendix~\ref{appendix:model-type-desc} and ~\ref{appendix:IFT_Setups_app}.

\noindent \textbf{Vocabulary Adaptation Baseline Models.} We obtain the \textit{BSA-PubMedBERT} baseline by replacing the encoder of BertSumAbs (BSA) with PubmedBERT~\cite{pubmedbert} respectively. In the same manner, we adapt a recent vocabulary adaptation model, AVocaDo~\cite{hong-etal-2021-avocado} from classification to the summarization setting. AVocaDo selects subwords from a vocabulary and iteratively builds on a downstream dataset until the fragment score stays above a particular threshold (taken as $3$).

\noindent \textbf{Intermediate Fine-tuning Baseline Models.}  BioBART~\cite{yuan-etal-2022-biobart} is obtained when BART is continuously pre-trained using PubMed abstracts corpora using only text-infilling as the pre-training objective. It achieves state-of-the-art performance on the MeQSum and CHQSum datasets. \textit{IFT-CNN} describes existing PLMs that only perform intermediate fine-tuning with the CNN/DailyMail dataset. Unlike \textit{IFT-CNN}, \textit{IFT-PAC} uses the PubMed Abstracts Collection (PAC) dataset for intermediate fine-tuning. \textit{IFT-PAC} is equivalent to \drift{} without vocabulary adaptation. \textit{BSA-BioBERT} is obtained similar to \textit{BSA-PubMedBERT} where the encoder of BertSumAbs (BSA) is replaced with BioBERT~\cite{cite:BioBert}.

\subsection{Training Details}\label{sec:pac-construct}
We obtain the biomedical paper abstracts from the official dump\footnote{\url{https://ftp.ncbi.nlm.nih.gov/pubmed/updatefiles}} of PubMed dated 2020. It comprises $450K$ data points (PubMed abstract and title). We refer to it as the \textit{PubMed Abstracts Collection (PAC)} in the paper. We remove data points that overlap with downstream datasets (as EBM and BioASQ contain PubMed abstracts as the source document), as a decontamination step to prevent memorization issues~\cite{radford2019language}. We randomly select $312K$ data points to form the final dataset. We keep the dataset size of PAC similar to that of CNN/DailyMail for a fair comparison. Appendix~\ref{appendix:training-details-app} provides further implementation details and the optimal hyperparameters and vocabulary size details.

\section{Experimental Results} \label{sec:result_discussion}
We show the performance comparison results of \drift{} in Table~\ref{tab:results_pipeline}. We observe an average Rouge-L improvement of $15.74\%$ across datasets over baselines in a zero-shot setting. We further observe gains of $10.81$\% and $17.29$\% across challenging scenarios of long-form medical summary generation and reference summaries with high OOV concentration respectively. The consistent improvement also holds for Concept Score that captures the faithfulness aspect~\cite{zhang2023famesumm}. \drift{} performs quite well for short-form summaries where it achieves SOTA performance on the MeQSum (even outperforming BioMedGPT~\cite{zhang2023biomedgpt}) and CHQSum data. %

\subsection{Performance Evaluation of \drift{}}\label{sec:disc-results}
We investigate the \drift{} performance using five research questions (RQs).

\noindent \textbf{RQ1: \drift{} outperforms vocabulary adaptation baselines.}
We observe that \drift{} outperforms the vocabulary adaptation baselines of BSA-PubMedBERT and BSA-AVocaDo by a good margin of $33.10\%$ and $3.94$\% respectively in terms of Rouge-L. We thus observe the effectiveness of designing the vocabulary adaptation as a hyperparameter tuning search (as explained in Section~\ref{sec:vocab-adapt-method}). We show consistent improvement due to the vocabulary adaptation step alone (\drift{} versus IFT-PAC) across three different PLMs in the case of EBM and BioASQ, where \drift{} improves over IFT-PAC by $3.55\%$ and $5.38\%$ respectively. We further observe that \drift{} outperforms proportionately to the percentage of OOV words (higher the OOV\%, \drift{} outperforms more). We thus observe limited improvement in the case of MeQSum and CHQSum where the OOV percentage is $11.63\%$ and $9.03\%$, which is much lower as compared to EBM and BioASQ where the OOV percentage is $14.67$\% and $18.46$\%. We extensively analyze the impact of the added vocabulary of \drift{} and AVoCaDo in terms of fragment score in the Appendix~\ref{appendix:eval-medvoc}. 

\begin{table}[t]
    \centering
    \footnotesize
    \scalebox{0.85}{
    \begin{tabular}{ccccc}
    \hline
              \textbf{Model}  & \textbf{MeQSum} & \textbf{CHQSum}& \textbf{EBM} & \textbf{BioASQ}\\ \hline
        BSA-PubMedBERT  &   39.79 &  30.59    & 17.76    & 26.65\\ 
        BSA-AVocaDo &  49.30 & 34.49 &  18.43 & 45.86\\ %
        \drift{}  (BSA)   &\best{51.49}    &  \best{35.11}    &\best{19.51}& \best{47.54}\\ \hline
    \end{tabular}}
    \caption{Rouge-L comparison with vocabulary adaptation baselines. \drift{} outperforms BSA-AVocaDo by $3.94$\% on average.}
    \label{tab:baseline-comp}
\end{table}

\begin{table*}[t]
    \centering
    \setlength{\tabcolsep}{0cm}
    \scalebox{0.65}{
    \begin{tabular}{cccccc|ccccc|ccccc|cc} \hline
     \textbf{Model}& \multicolumn{5}{c}{\textbf{BertSumAbs (BSA)}}& \multicolumn{5}{c}{\textbf{BART}}& \multicolumn{5}{c}{\textbf{PEGASUS}}& \multicolumn{2}{c}{\textbf{Overall}}\\
     &\textbf{R-1}&\textbf{R-2}&\textbf{R-L}&\textbf{BSr}&\textbf{CSr}&\textbf{R-1}&\textbf{R-2}&\textbf{R-L}&\textbf{BSr}&\textbf{CSr}&\textbf{R-1}&\textbf{R-2}&\textbf{R-L}&\textbf{BSr}&\textbf{CSr}&\textbf{R-L}&\textbf{CSr}\\ \hline
     \multicolumn{18}{c}{\textbf{EBM}} \\ 
     SOTA&25.40&7.06&18.88&85.19&18.44&\best{29.41}&\best{9.15}&\secondbest{20.62}&\secondbest{85.97}&\best{24.14}&25.46&6.82&18.09&85.75&20.31&19.19&20.96\\

     IFT-CNN&26.37&6.37&18.79&84.76&18.14&27.06&7.66&19.08&85.76&21.72&25.46&6.82&18.09&\best{85.75}&20.31&18.65&20.06\\
     
     IFT-PAC&\secondbest{27.22}&\secondbest{7.53}&\secondbest{19.26}&\secondbest{84.79}&\secondbest{19.58}&28.30&7.98&19.78&85.84&21.82&\secondbest{26.69}&\secondbest{7.93}&\secondbest{19.01}&85.47&\best{22.78}&\secondbest{19.35}&\secondbest{21.39}\\
     
     \drift{}&\best{27.67}&\best{8.01}&\best{19.51}&\best{85.05}&\best{20.36}&\secondbest{29.22}&\secondbest{8.62}&\best{20.65}&\best{86.17}&\secondbest{22.66}&\best{29.12}&\best{8.41}&\best{19.95}&\secondbest{85.71}&\secondbest{22.41}&\best{20.03}&\best{21.81} \\  %

    \hline
    \multicolumn{18}{c}{{\bf BioASQ}} \\ 
    
    SOTA&49.05&37.15&\secondbest{45.84}&\best{{90.55}}&50.57&\secondbest{51.78}&\best{{39.91}}&\secondbest{47.36}& \secondbest{90.77} & \secondbest{52.05} &44.63&29.77&39.48 & 89.44 & \best{{47.82}} & \secondbest{44.23}&\secondbest{50.14}\\

     IFT-CNN& 45.65&33.48&42.17 & 89.27 & 44.61 & 48.84&37.41&45.29 & 90.31 &  49.48 & 44.63&29.77&39.48 & 89.44  & \best{{47.82}}&  42.31 & 47.30\\
     
     IFT-PAC & \secondbest{49.58} & \secondbest{37.82} &44.76 & 89.81 & \secondbest{50.96} & 50.32&38.26&45.00 & 90.45 & 49.53 &\secondbest{45.37}&\best{{34.80}}& \secondbest{41.06} & \secondbest{89.70}  & \secondbest{46.85}& 43.60  & 49.02\\
     
      \drift{}& \best{{52.03}}& \best{40.44} & \best{{47.54}} & \secondbest{90.48} & \best{{52.20}} & \best{{52.48}} &\secondbest{39.16}& \best{{48.02}} & \best{{91.16}} &  \best{{52.87}} & \best{{47.44}} &\secondbest{33.49}& \best{{42.39}} & \best{{89.94}} & 46.42&  \best{{45.98}} & \best{{50.50}}\\  %
     
    \hline
    \multicolumn{18}{c}{{\bf MeQSum}} \\ 
    
     SOTA& \secondbest{52.64} & \secondbest{37.66} & \secondbest{49.99} & \secondbest{93.55} & \best{{53.56}} & 55.53 & 40.31 & 52.67 & 93.99 & 58.10 & 53.87&38.65&51.03 & 93.88 & 55.84 & 51.23 & 55.83 \\

     IFT-CNN&46.92&30.53&44.33 & 91.72 & 47.48& \best{{59.49}} & \secondbest{43.24} & \best{{56.16}} & \best{{94.83}} & \secondbest{60.65} & 53.87 & 38.65 & 51.03 & \secondbest{93.88} & 55.84 &  50.51& 54.63\\
    
     IFT-PAC& 49.44&33.31&46.28 & 92.89 & 49.76 & \secondbest{59.09} & 42.76 & 55.73 & 93.90  & \best{{61.30}} & \best{{58.24}} & \best{{43.45}} & \best{{55.39}} &  \best{{94.31}}  & \best{{61.07}} &  \secondbest{52.47} &  \secondbest{57.38}\\
     
     \drift{}  & \best{{54.65}}& \best{{38.70}}& \best{{51.49}}& \best{{93.62}} & \secondbest{53.44} & 58.44& \best{{44.40}}&\secondbest{55.88} & \secondbest{94.20} & 60.52 & \secondbest{56.30} & \secondbest{40.86} & \secondbest{53.52} & 93.18  & \secondbest{59.25} &  \best{{53.63}} &   \best{{57.73}} \\  %
    \hline
    \multicolumn{18}{c}{{\bf CHQSum}} \\
    SOTA& 35.99 & 16.96 & 33.72 & 91.01 & 33.45 & 40.44 & 21.04 & 38.51 & 91.98 & 38.73 & 43.07&24.11&40.44 & 92.04 & 42.33 & 37.56 & 38.17\\
    
    IFT-CNN& 37.81&\best{19.14}&\secondbest{35.64}& 91.02 & \best{34.82} & \secondbest{41.07}&\secondbest{22.18}&\secondbest{39.02} & \best{92.09} & \secondbest{42.19} &\secondbest{43.07}&\best{24.11}&\secondbest{40.44} & \best{92.04}  &\secondbest{42.33} & 38.36& \secondbest{39.78}\\
    
    IFT-PAC&\best{38.58}&\secondbest{18.56}&\best{36.24} & \best{91.36}  & \secondbest{34.28} & 40.53&21.16&38.75 &  91.92&   41.02& 42.73&23.55&40.35 & 91.92 & 41.44 &  \secondbest{38.45}& 38.91\\ 
    
    \drift{}&\secondbest{37.87}&18.25& 35.11 & \secondbest{91.10} & 33.56 & \best{42.58}&\best{24.02}&\best{40.59} & \secondbest{92.05} & \best{45.63} & \best{43.10} & \secondbest{24.09} & \best{40.57} & \secondbest{92.02} & \best{43.43} & \best{38.75} & \best{40.84}\\  \hline
    \end{tabular}}
    \caption{Performance comparison of \drift{} with Rouge-L (\textbf{R-L}) as the primary metric; we highlight the \best{best} and \secondbest{second-best} settings. IFT-PAC is equivalent to \drift{} without vocabulary adaptation. The improvements wherever observed in \drift{} over IFT-CNN for \textbf{R-L} are \textbf{statistically significant} across all settings (using paired t-test; $p <0.01$). \drift{} generates \textbf{more informative} (improves overall R-L by $5.99\%$ on average) and \textbf{more faithful} medical summaries (improves overall Concept Score by $5.96\%$ on average). For SOTA, we use BioBERT, BioBART, and IFT-CNN for BertSumAbs, BART, and PEGASUS respectively.}
    \label{tab:results_pipeline}
\end{table*}

\noindent \textbf{RQ2: \drift{} outperforms baselines even in zero and few-shot summarization tasks.} We observe from Figure~\ref{fig:x-shot-eval}(a) that \drift{} consistently outperforms IFT-CNN across the full dataset, leading to improved zero-shot and few-shot ($10$ and $100$-shot) abstractive summarization for CHQSum, MeQSum and EBM, with average performance gains of $28.94$\% and $8.13$\% in terms of Rouge-L, respectively. Remarkably, the advantage of \drift{} is more pronounced in zero and few-shot settings compared to training on the entire dataset ($15.74$\% versus $5.99$\%). In contrast, for the BioASQ dataset, IFT-CNN exhibits higher zero-shot performance than \drift{} (Rouge-L score of $35.36$), increasing to $42.31$ in the full data setting (\drift{} outperforms in full data setting by $8.67$\%) as reference summaries of BioASQ is extractive (characterized by unigram and bigram overlaps between SD and RS of $96.72\%$ and $84.30\%$, respectively), similar to the well-known extractive nature~\cite{liulapata2019text} of CNN/DailyMail dataset. 

\begin{figure}[!ht]
    \centering
    \subfigure[]{\includegraphics[scale=0.37]{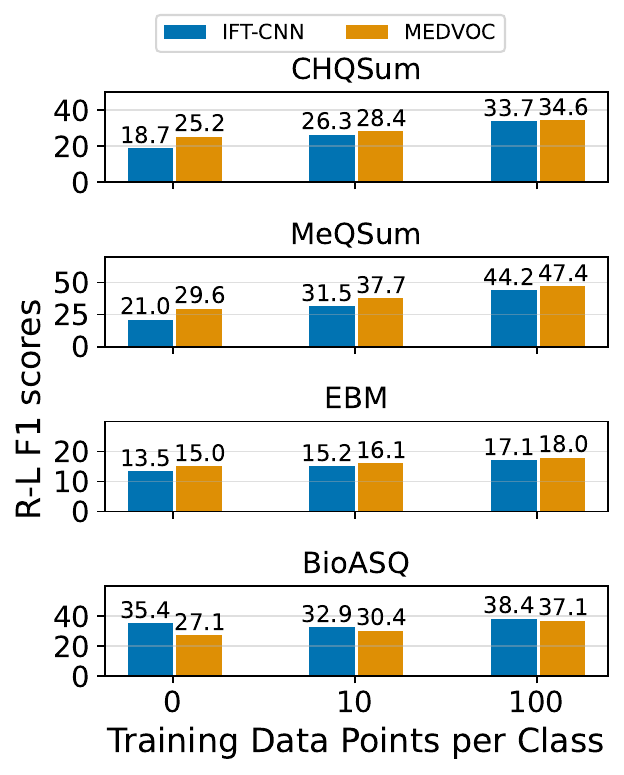}}
    \subfigure[]{\includegraphics[scale=0.4]{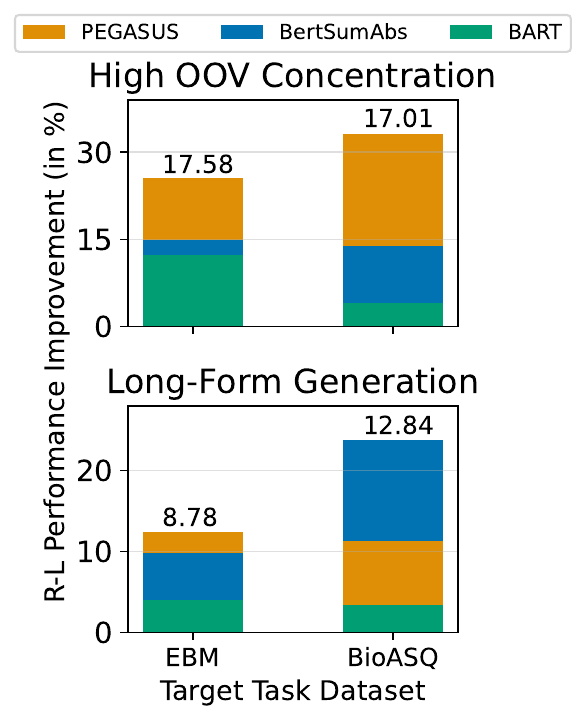}}
    \caption{(a) Zero-shot and few-shot performance in terms of Rouge-L scores averaged across the three PLMs. \drift{} shows statistically significant improvement over IFT-CNN in most settings, except BioASQ. (b) Performance improvement of \drift{} over IFT-CNN in high OOV concentration and long-form reference summaries (top-ten percentile); PEGASUS shows the highest jump.} %
    \label{fig:x-shot-eval}
\end{figure}

\noindent \textbf{RQ3: \drift{} outperforms baselines even when reference summaries have high OOV concentration.}
We select the top ten percentile of data points that have the highest OOV concentration in reference summaries. These points represent the most difficult data points in terms of vocabulary mismatch. \drift{} shows a high improvement of $17.29$\% on average over IFT-CNN in the BioASQ and EBM datasets that have an average OOV concentration of $41.95\%$ in the selected data points. Figure~\ref{fig:x-shot-eval}(b) shows that higher the OOV concentration, higher the performance improvement as the PEGASUS model type with BioASQ dataset shows the highest performance jump of $40.53$\% over IFT-CNN. %

\noindent \textbf{RQ4: \drift{} outperforms baselines in case of longer reference summaries.} Generating long-form medical summaries is under-explored~\cite{liu2023task}. Here, we select data points where RS length is greater than $30$ tokens (which is the $95$ percentile of length of summaries for MeQSum and CHQSum) and limit our evaluation to EBM and BioASQ. We observe that across these two datasets and three model types, \drift{} outperforms IFT-CNN by $10.81$\% on average, which is quite higher than $8.23\%$ improvement on the entire dataset. Therefore, \drift{} improves the generation of long-form medical summaries (Figure \ref{fig:x-shot-eval}(b)).

\noindent \textbf{RQ5: \drift{} trains the model decoder to incorporate more relevant medical words during generation and produce more faithful medical summaries.} We observe that BertSumAbs shows the highest performance improvement (based on Rouge-L) on average of $6.26\%$ due to the vocabulary addition step, among the three PLMs. This is because the number of subwords required to generate a medical concept-bearing word (fragment score) significantly reduced from $2.09$ to $1.58$, the highest drop of $25\%$ as compared to BART and PEGASUS of $3.47\%$ and $3.14\%$ respectively. We further observe that the \textit{faithfulness} aspect measured using \textit{Concept Score} (CSr) in a medical context~\cite{zhang2023famesumm} improved significantly on average across EBM and BioASQ, by $8.72\%$, $6.76\%$ respectively. This indicates that \drift{} generates more faithful summaries. Furthermore, we observe that a large percentage of top-3 candidate beams (in terms of candidate beam score computed as the average \textit{negative log likelihood} scores of the tokens in the beam) in the \textit{BertSumAbs} model type contains $85.92\%$, $85.37\%$, and $86.16\%$ of newly added \drift{} vocabulary in the case of first, second, and third candidate beam respectively. This further highlights the positive impact of vocabulary adaptation for medical abstractive summarization.

\subsection{Ablation Analysis} \label{sec:ablation-analysis}
We observe an average Rouge-L improvement of $29\%$ for IFT-PAC when compared with the IFT-CNN model in the case of EBM and BioASQ, in the high OOV concentration in the RS setting (as done in RQ3), where PEGASUS model on BioASQ dataset shows the highest performance improvement of $64.51\%$, as also observed in RQ3. %

\noindent \textbf{Biomedical paper title generations serve as a good intermediate fine-tuning task for medical summarization.} We observe the following characteristics of PAC that are similar to the downstream datasets. First, the source document length of EBM and BioASQ, of $276$ and $505$ words, is more similar to the PAC dataset ($276$ words) as compared to CNN/DailyMail ($700$ words). Second, in the case of MeQSum and CHQSum, the length of RS is $12$ words on average, and it is almost the same as PAC ($15$ words) as compared to CNN/DailyMail ($57$ words). Third, the abstractive nature of summaries of PAC aligns more with the downstream datasets. PAC, EBM, and MeQSum have a bigram overlap between SD and RS of $33.33\%$, $15.39\%$ and $10.10\%$, respectively. 

\noindent \textbf{Biomedical paper title generation successfully aligns the positional embeddings to the target (medical) domain.} The most informative part of a PubMed abstract is predominantly located in the \textit{Conclusion} and \textit{Results} section~\cite{DBLP:conf/acl-alta/MollaS11a,jin-etal-2019-pubmedqa}, which naturally makes these sections more probable and desirable to be a part of a reference summary. Therefore, we observe in Figure~\ref{fig:shift-positional}(a) that the positional embedding shift due to IFT-CNN over occurs the most at the start, whereas a consistently higher shift is seen for IFT-PAC after the initial around fifty tokens and the highest shift occurs towards the end of a SD of a PubMed abstract (token index $\geq 400$). Thus, the higher domain mismatch in the case of the biomedical literature domain is well-captured by a higher positional embedding shift. %

\begin{figure}[t]
    \centering
    \subfigure[]{\includegraphics[width=0.22\textwidth]{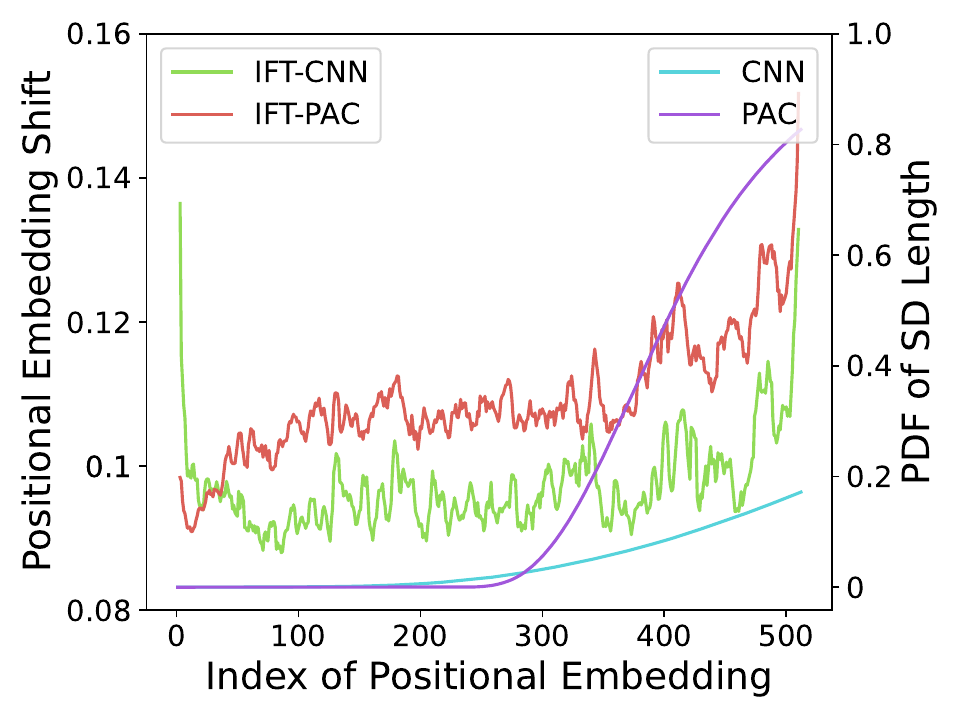}}
    \subfigure[]{\includegraphics[width=0.22\textwidth]{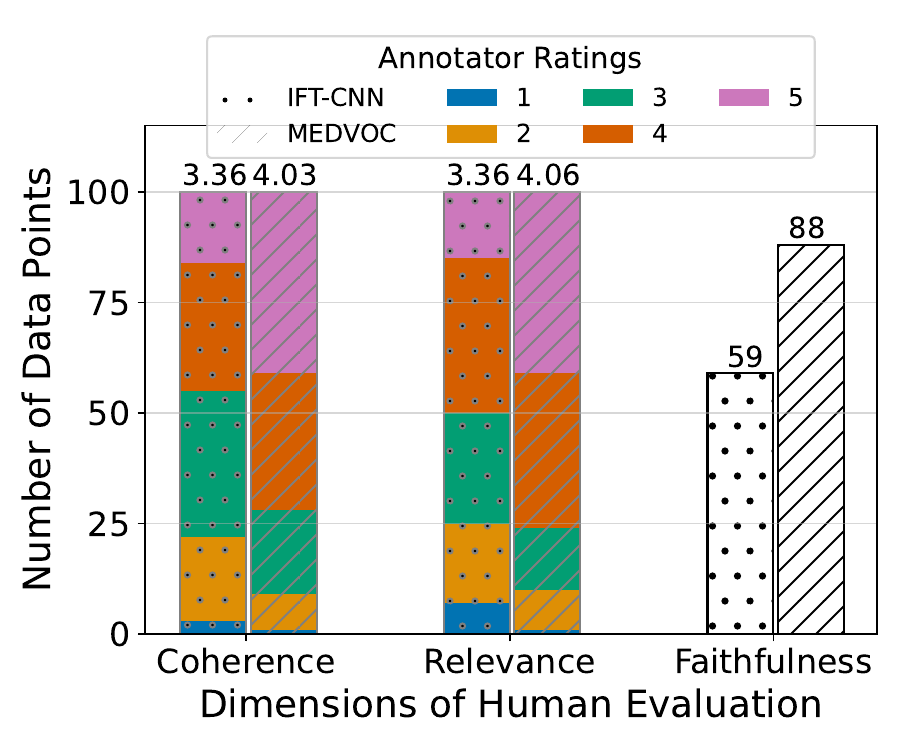}}
    \caption{(a) Shift observed in the positional embedding for BertSumAbs in terms of Euclidean distance. (b) Human evaluation scores comparison over $100$ randomly selected test data points. \drift{} produces more relevant, coherent, and faithful summaries during human evaluation with medical experts.} %
    \label{fig:shift-positional}
\end{figure}

\subsection{Human Evaluation}\label{sec:human-evaluation}
We randomly select $100$ test data points uniformly across the four datasets and use the Prolific platform to recruit medical experts (self-reported to have a Masters or Doctoral degree in Medicine and Biomedical Sciences and are older than $24$ years) for annotating summary pairs of \drift{} and IFT-CNN across the standard aspects~\cite{fabbri-etal-2021-summeval,zhang2023famesumm} of \textit{relevance}, \textit{coherence} (on a Likert scale of $1$ to $5$), and faithfulness (binary). Each annotator was given $30$ minutes to evaluate $10$ summaries and was compensated at a rate of $9$ UK pounds per hour (see Appendix~\ref{appendix:human-eval-app} for more details). Figure~\ref{fig:shift-positional}(b) shows the human evaluation results where \drift{} generates more faithful summaries ($88\%$ versus $59\%$ of summaries are faithful), and more relevant summaries, where $76\%$ of data points get a positive score of $4$ or $5$ in Likert scale, as compared to $50\%$ by IFT-CNN.

\section{Conclusion}\label{sec:conclusion}
We present a dynamic vocabulary adaptation strategy, called \drift, for fine-tuning PLMs for improved medical text summarization. To the best of our knowledge, this is the first work that uses vocabulary adaptation techniques for summarization and opens up an interesting potential research direction. Through extensive experimentation, we observe that \drift{} consistently outperforms vocabulary adaptation baselines and significantly outperforms standard fine-tuning strategy (IFT-CNN) in full data setting. \drift{} outperforms even in zero and few-shot settings, as well as when reference summaries have high OOV concentration or are long ($> 30$ tokens). \drift{} outperforms IFT-CNN by a high margin in terms of relevance and faithfulness in human evaluation with medical experts. As an immediate future work, we will extend to the multi-document summarization setting. Given that \drift{} leads to more faithful summaries, we will incorporate vocabulary adaptation to improve state-of-the-art models~\cite{alambo2022improving,zhang2023famesumm,xie2023factreranker} that improve the factual consistency of summaries.  %

\section*{Acknowledgement}
Gunjan Balde is supported by the Prime Minister Research Fellowship (PMRF) in India through grant number IIT/Acad/PMRF/Autumn/2021-22 (Application Number: PMRF-192002-1521) dated 18 November 2021. Soumyadeep Roy is supported by the Institute Ph.D. Fellowship at the Indian Institute of Technology Kharagpur. This research was partially funded by the Federal Ministry of Education and Research (BMBF), Germany under the project LeibnizKILabor with grant No. 01DD20003 and by a Google India Faculty Research Award.
\bibliographystyle{named}
\bibliography{custom}

\begin{thebibliography}{}

\bibitem[\protect\citeauthoryear{Alambo \bgroup \em et al.\egroup
  }{2022}]{alambo2022improving}
Amanuel Alambo, Tanvi Banerjee, et~al.
\newblock Improving the factual accuracy of abstractive clinical text
  summarization using multi-objective optimization.
\newblock In {\em 2022 44th Annual International Conference of the IEEE
  Engineering in Medicine \& Biology Society (EMBC)}, pages 1615--1618. IEEE,
  2022.

\bibitem[\protect\citeauthoryear{Allison \bgroup \em et al.\egroup
  }{2006}]{allison2006another}
Ben Allison, David Guthrie, et~al.
\newblock Another look at the data sparsity problem.
\newblock In {\em Text, Speech and Dialogue: 9th International Conference, TSD
  2006, Brno, Czech Republic, September 11-15, 2006. Proceedings 9}, pages
  327--334. Springer, 2006.

\bibitem[\protect\citeauthoryear{Ben~Abacha and
  Demner-Fushman}{2019}]{ben-abacha-demner-fushman-2019-summarization}
Asma Ben~Abacha and Dina Demner-Fushman.
\newblock On the summarization of consumer health questions.
\newblock In {\em Proceedings of the 57th Annual Meeting of the Association for
  Computational Linguistics}, pages 2228--2234, July 2019.

\bibitem[\protect\citeauthoryear{Chang and
  Lu}{2021}]{chang-lu-2021-rethinking-intermediate}
Ting-Yun Chang and Chi-Jen Lu.
\newblock Rethinking why intermediate-task fine-tuning works.
\newblock In {\em Findings of the Association for Computational Linguistics:
  EMNLP 2021}, pages 706--713, November 2021.

\bibitem[\protect\citeauthoryear{Chen \bgroup \em et al.\egroup
  }{2023}]{chen2023meditron}
Zeming Chen, Alejandro~Hern{\'a}ndez Cano, et~al.
\newblock Meditron-70b: Scaling medical pretraining for large language models.
\newblock {\em arXiv preprint arXiv:2311.16079}, 2023.

\bibitem[\protect\citeauthoryear{Dai \bgroup \em et al.\egroup
  }{2021}]{dai-etal-2021-bdkg}
Songtai Dai, Quan Wang, et~al.
\newblock {BDKG} at {MEDIQA} 2021: System report for the radiology report
  summarization task.
\newblock In {\em Proceedings of the 20th Workshop on Biomedical Language
  Processing}, pages 103--111, June 2021.

\bibitem[\protect\citeauthoryear{Diao \bgroup \em et al.\egroup
  }{2021}]{diao-etal-2021-taming}
Shizhe Diao, Ruijia Xu, et~al.
\newblock Taming pre-trained language models with n-gram representations for
  low-resource domain adaptation.
\newblock In {\em Proceedings of the 59th Annual Meeting of the Association for
  Computational Linguistics and the 11th International Joint Conference on
  Natural Language Processing (Volume 1: Long Papers)}, pages 3336--3349,
  August 2021.

\bibitem[\protect\citeauthoryear{Fabbri \bgroup \em et al.\egroup
  }{2021a}]{fabbri-etal-2021-improving}
Alexander Fabbri, Simeng Han, et~al.
\newblock Improving zero and few-shot abstractive summarization with
  intermediate fine-tuning and data augmentation.
\newblock In {\em Proceedings of the 2021 Conference of the North American
  Chapter of the Association for Computational Linguistics: Human Language
  Technologies}, pages 704--717, June 2021.

\bibitem[\protect\citeauthoryear{Fabbri \bgroup \em et al.\egroup
  }{2021b}]{fabbri-etal-2021-summeval}
Alexander~R. Fabbri, Wojciech Kry{\'s}ci{\'n}ski, et~al.
\newblock {S}umm{E}val: Re-evaluating summarization evaluation.
\newblock {\em Transactions of the Association for Computational Linguistics},
  9:391--409, 2021.

\bibitem[\protect\citeauthoryear{Gu \bgroup \em et al.\egroup
  }{2021}]{pubmedbert}
Yu~Gu, Robert Tinn, Cheng, et~al.
\newblock Domain-specific language model pretraining for biomedical natural
  language processing.
\newblock {\em ACM Trans. Comput. Healthcare}, 3(1), oct 2021.

\bibitem[\protect\citeauthoryear{He \bgroup \em et al.\egroup
  }{2021}]{he-etal-2021-damo}
Yifan He, Mosha Chen, et~al.
\newblock damo{\_}nlp at {MEDIQA} 2021: Knowledge-based preprocessing and
  coverage-oriented reranking for medical question summarization.
\newblock In {\em Proceedings of the 20th Workshop on Biomedical Language
  Processing}, pages 112--118, June 2021.

\bibitem[\protect\citeauthoryear{Hong \bgroup \em et al.\egroup
  }{2021}]{hong-etal-2021-avocado}
Jimin Hong, TaeHee Kim, et~al.
\newblock {AV}oca{D}o: Strategy for adapting vocabulary to downstream domain.
\newblock In {\em Proceedings of the 2021 Conference on Empirical Methods in
  Natural Language Processing}, pages 4692--4700, November 2021.

\bibitem[\protect\citeauthoryear{Jin \bgroup \em et al.\egroup
  }{2019}]{jin-etal-2019-pubmedqa}
Qiao Jin, Bhuwan Dhingra, et~al.
\newblock {P}ub{M}ed{QA}: A dataset for biomedical research question answering.
\newblock In {\em Proceedings of the EMNLP-IJCNLP 2019}, pages 2567--2577,
  November 2019.

\bibitem[\protect\citeauthoryear{Kanwal and Rizzo}{2022}]{kunwal-rizzo-et-al}
Neel Kanwal and Giuseppe Rizzo.
\newblock Attention-based clinical note summarization.
\newblock In {\em Proceedings of the 37th ACM/SIGAPP Symposium on Applied
  Computing}, SAC '22, page 813–820, 2022.

\bibitem[\protect\citeauthoryear{Lamproudis. \bgroup \em et al.\egroup
  }{2022}]{healthinf22}
Anastasios Lamproudis., Aron Henriksson., et~al.
\newblock Vocabulary modifications for domain-adaptive pretraining of clinical
  language models.
\newblock In {\em Proceedings of the 15th International Joint Conference on
  Biomedical Engineering Systems and Technologies (BIOSTEC 2022) - HEALTHINF},
  pages 180--188. INSTICC, 2022.

\bibitem[\protect\citeauthoryear{Laskar \bgroup \em et al.\egroup
  }{2022}]{laskar2022}
Md~Tahmid~Rahman Laskar, Enamul Hoque, et~al.
\newblock {Domain Adaptation with Pre-trained Transformers for Query-Focused
  Abstractive Text Summarization}.
\newblock {\em Computational Linguistics}, 48(2):279--320, 06 2022.

\bibitem[\protect\citeauthoryear{Lee \bgroup \em et al.\egroup
  }{2019}]{cite:BioBert}
Jinhyuk Lee, Wonjin Yoon, et~al.
\newblock {BioBERT: a pre-trained biomedical language representation model for
  biomedical text mining}.
\newblock {\em Bioinformatics}, 36(4):1234--1240, 09 2019.

\bibitem[\protect\citeauthoryear{Lewis \bgroup \em et al.\egroup
  }{2020}]{lewis2020bart}
Mike Lewis, Yinhan Liu, et~al.
\newblock Bart: Denoising sequence-to-sequence pre-training for natural
  language generation, translation, and comprehension.
\newblock In {\em Proceedings of the 58th Annual Meeting of the Association for
  Computational Linguistics}, pages 7871--7880, 2020.

\bibitem[\protect\citeauthoryear{Lin}{2004}]{lin-2004-rouge}
Chin-Yew Lin.
\newblock {ROUGE}: A package for automatic evaluation of summaries.
\newblock In {\em Text Summarization Branches Out}, pages 74--81, 2004.

\bibitem[\protect\citeauthoryear{Liu and Lapata}{2019}]{liulapata2019text}
Yang Liu and Mirella Lapata.
\newblock Text summarization with pretrained encoders.
\newblock In {\em Proceedings of EMNLP-IJCNLP}, pages 3730--3740, 2019.

\bibitem[\protect\citeauthoryear{Liu \bgroup \em et al.\egroup
  }{2023}]{liu2023task}
Siyang Liu, Naihao Deng, et~al.
\newblock Task-adaptive tokenization: Enhancing long-form text generation
  efficacy in mental health and beyond.
\newblock In {\em Proceedings of the 2023 Conference on Empirical Methods in
  Natural Language Processing}, pages 15264--15281, 2023.

\bibitem[\protect\citeauthoryear{Moll{\'{a}} and
  Santiago{-}Mart{\'{\i}}nez}{2011}]{DBLP:conf/acl-alta/MollaS11a}
Diego Moll{\'{a}} and Mar{\'{\i}}a~Elena Santiago{-}Mart{\'{\i}}nez.
\newblock Development of a corpus for evidence based medicine summarisation.
\newblock In {\em Proceedings of the Australasian Language Technology
  Association Workshop}, pages 86--94, 2011.

\bibitem[\protect\citeauthoryear{Paul \bgroup \em et al.\egroup
  }{2022}]{spaul_legal}
Shounak Paul, Arpan Mandal, et~al.
\newblock Pre-training transformers on indian legal text, 2022.

\bibitem[\protect\citeauthoryear{Phang \bgroup \em et al.\egroup
  }{2019}]{phang2019sentence}
Jason Phang, Thibault Févry, et~al.
\newblock Sentence encoders on stilts: Supplementary training on intermediate
  labeled-data tasks, 2019.

\bibitem[\protect\citeauthoryear{Radford \bgroup \em et al.\egroup
  }{2019}]{radford2019language}
Alec Radford, Jeffrey Wu, et~al.
\newblock Language models are unsupervised multitask learners.
\newblock {\em OpenAI blog}, 1(8):9, 2019.

\bibitem[\protect\citeauthoryear{Rust \bgroup \em et al.\egroup
  }{2021}]{rust-etal-2021-good}
Phillip Rust, Jonas Pfeiffer, et~al.
\newblock How good is your tokenizer? on the monolingual performance of
  multilingual language models.
\newblock In {\em Proceedings of the 59th Annual Meeting of the Association for
  Computational Linguistics and the 11th International Joint Conference on
  Natural Language Processing (Volume 1: Long Papers)}, pages 3118--3135,
  August 2021.

\bibitem[\protect\citeauthoryear{Schick and
  Schütze}{2020}]{Schick_Schütze_2020}
Timo Schick and Hinrich Schütze.
\newblock Rare words: A major problem for contextualized embeddings and how to
  fix it by attentive mimicking.
\newblock {\em Proceedings of the AAAI Conference on Artificial Intelligence},
  34(05):8766--8774, Apr. 2020.

\bibitem[\protect\citeauthoryear{See \bgroup \em et al.\egroup
  }{2017}]{DBLP:conf/acl/SeeLM17}
Abigail See, Peter~J. Liu, et~al.
\newblock Get to the point: Summarization with pointer-generator networks.
\newblock In {\em Proceedings of the 55th Annual Meeting of the Association for
  Computational Linguistics, Volume 1: Long Papers}, pages 1073--1083, 2017.

\bibitem[\protect\citeauthoryear{Soldaini and
  Goharian}{2016}]{soldaini2016quickumls}
Luca Soldaini and Nazli Goharian.
\newblock Quickumls: a fast, unsupervised approach for medical concept
  extraction.
\newblock In {\em MedIR workshop, SIGIR}, pages 1--4, 2016.

\bibitem[\protect\citeauthoryear{Suresh \bgroup \em et al.\egroup
  }{2023}]{suresh-etal-2023-intermediate}
Shilpa Suresh, Nazgol Tavabi, et~al.
\newblock Intermediate domain finetuning for weakly supervised domain-adaptive
  clinical {NER}.
\newblock In {\em The 22nd Workshop on Biomedical Natural Language Processing
  and BioNLP Shared Tasks}, pages 320--325, July 2023.

\bibitem[\protect\citeauthoryear{Tai \bgroup \em et al.\egroup
  }{2020}]{tai-etal-2020-exbert}
Wen Tai, H.~T. Kung, et~al.
\newblock ex{BERT}: Extending pre-trained models with domain-specific
  vocabulary under constrained training resources.
\newblock In {\em Findings of the Association for Computational Linguistics:
  EMNLP 2020}, pages 1433--1439, November 2020.

\bibitem[\protect\citeauthoryear{Tsatsaronis \bgroup \em et al.\egroup
  }{2015}]{cite:BioASQ}
George Tsatsaronis, Georgios Balikas, et~al.
\newblock An overview of the bioasq large-scale biomedical semantic indexing
  and question answering competition.
\newblock {\em BMC Bioinformatics}, 16:138, 2015.

\bibitem[\protect\citeauthoryear{Wu \bgroup \em et al.\egroup
  }{2016}]{wu2016googles}
Yonghui Wu, Mike Schuster, et~al.
\newblock Google's neural machine translation system: Bridging the gap between
  human and machine translation, 2016.

\bibitem[\protect\citeauthoryear{Wu \bgroup \em et al.\egroup
  }{2023}]{wu2023pmcllama}
Chaoyi Wu, Weixiong Lin, et~al.
\newblock Pmc-llama: Towards building open-source language models for medicine,
  2023.

\bibitem[\protect\citeauthoryear{Xie \bgroup \em et al.\egroup
  }{2022}]{xie2022pre}
Qianqian Xie, Jennifer~Amy Bishop, et~al.
\newblock Pre-trained language models with domain knowledge for biomedical
  extractive summarization.
\newblock {\em Knowledge-Based Systems}, 252:109460, 2022.

\bibitem[\protect\citeauthoryear{Xie \bgroup \em et al.\egroup
  }{2023a}]{xie2023factreranker}
Qianqian Xie, Jinpeng Hu, et~al.
\newblock Factreranker: Fact-guided reranker for faithful radiology report
  summarization.
\newblock {\em arXiv preprint arXiv:2303.08335}, 2023.

\bibitem[\protect\citeauthoryear{Xie \bgroup \em et al.\egroup
  }{2023b}]{xie2023survey}
Qianqian Xie, Zheheng Luo, et~al.
\newblock A survey for biomedical text summarization: From pre-trained to large
  language models, 2023.

\bibitem[\protect\citeauthoryear{Xu \bgroup \em et al.\egroup
  }{2021}]{xu-etal-2021-vocabulary}
Jingjing Xu, Hao Zhou, et~al.
\newblock Vocabulary learning via optimal transport for neural machine
  translation.
\newblock In {\em Proceedings of the 59th Annual Meeting of the Association for
  Computational Linguistics and the 11th International Joint Conference on
  Natural Language Processing (Volume 1: Long Papers)}, pages 7361--7373,
  August 2021.

\bibitem[\protect\citeauthoryear{Xu \bgroup \em et al.\egroup
  }{2023}]{xu-etal-2023-retrieval}
Benfeng Xu, Chunxu Zhao, et~al.
\newblock Retrieval-augmented domain adaptation of language models.
\newblock In {\em Proceedings of the 8th Workshop on Representation Learning
  for NLP (RepL4NLP 2023)}, pages 54--64, July 2023.

\bibitem[\protect\citeauthoryear{Yadav \bgroup \em et al.\egroup
  }{2022}]{yadav2022chq}
Shweta Yadav, Deepak Gupta, et~al.
\newblock Chq-summ: A dataset for consumer healthcare question summarization.
\newblock {\em arXiv preprint arXiv:2206.06581}, 2022.

\bibitem[\protect\citeauthoryear{Yuan \bgroup \em et al.\egroup
  }{2022}]{yuan-etal-2022-biobart}
Hongyi Yuan, Zheng Yuan, et~al.
\newblock {B}io{BART}: Pretraining and evaluation of a biomedical generative
  language model.
\newblock In {\em Proceedings of the 21st Workshop on Biomedical Language
  Processing}, pages 97--109, May 2022.

\bibitem[\protect\citeauthoryear{Zhang \bgroup \em et al.\egroup
  }{2020a}]{pmlr-v119-zhang20ae}
Jingqing Zhang, Yao Zhao, et~al.
\newblock {PEGASUS}: Pre-training with extracted gap-sentences for abstractive
  summarization.
\newblock In {\em Proceedings of the 37th International Conference on Machine
  Learning}, volume 119 of {\em Proceedings of Machine Learning Research},
  pages 11328--11339, 13--18 Jul 2020.

\bibitem[\protect\citeauthoryear{Zhang \bgroup \em et al.\egroup
  }{2020b}]{ZhangKWWA20}
Tianyi Zhang, Varsha Kishore, et~al.
\newblock Bertscore: Evaluating text generation with {BERT}.
\newblock In {\em 8th International Conference on Learning Representations,
  {ICLR}}, 2020.

\bibitem[\protect\citeauthoryear{Zhang \bgroup \em et al.\egroup
  }{2021}]{zhang-etal-2021-leveraging-pretrained}
Longxiang Zhang, Renato Negrinho, et~al.
\newblock Leveraging pretrained models for automatic summarization of
  doctor-patient conversations.
\newblock In {\em Findings of the Association for Computational Linguistics:
  EMNLP 2021}, pages 3693--3712, November 2021.

\bibitem[\protect\citeauthoryear{Zhang \bgroup \em et al.\egroup
  }{2023a}]{zhang2023biomedgpt}
Kai Zhang, Jun Yu, et~al.
\newblock Biomedgpt: A unified and generalist biomedical generative pre-trained
  transformer for vision, language, and multimodal tasks.
\newblock {\em arXiv preprint arXiv:2305.17100}, 2023.

\bibitem[\protect\citeauthoryear{Zhang \bgroup \em et al.\egroup
  }{2023b}]{zhang2023famesumm}
Nan Zhang, Yusen Zhang, et~al.
\newblock Famesumm: Investigating and improving faithfulness of medical
  summarization.
\newblock {\em arXiv preprint arXiv:2311.02271}, 2023.

\bibitem[\protect\citeauthoryear{Zhu \bgroup \em et al.\egroup
  }{2023}]{zhu2023parameterefficient}
Yunqi Zhu, Xuebing Yang, et~al.
\newblock Parameter-efficient fine-tuning with layer pruning on free-text
  sequence-to-sequence modeling, 2023.

\end{thebibliography}

\appendix

\section{Experimental Setup}\label{appendix:expt-setup}

\subsection{Pre-trained Language Models}\label{appendix:model-type-desc}
To test the generalizability of our method described in Section~\ref{sec:methods}, we evaluate the efficacy of \drift{} on three State-of-the-art encoder-decoder-based PLMs. %

\begin{itemize}
    \item {\bf BertSumAbs} (BSA)~\cite{liulapata2019text}: It uses a standard encoder-decoder framework where Bert acts as an encoder, and the decoder is six-layered transformer architecture initialized randomly\footnote{\url{https://github.com/nlpyang/PreSumm}}. BertSumAbs has $180$ Million parameters, uses the \textit{Word-Piece} tokenizer and its pretraining objective is a combination of \textit{Masked Language Modeling} and \textit{Next Sentence Prediction}. The vocabulary size of this PLM ($|V_{\text{PLM}}|$) is $30522$. %

    \item \textbf{BART}~\cite{lewis2020bart}: BART is a denoising autoencoder, implemented as a sequence-to-sequence model with a bidirectional encoder over corrupted text and a left-to-right auto-regressive decoder to generate the original document it was derived from. We use the BART-LARGE\footnote{\url{https://huggingface.co/facebook/bart-large}} model available from the \textit{huggingface} library. BART has $406$ Million parameters, uses \textit{Byte-Pair Encoding} tokenization, and its pretraining objective is a combination of \textit{Text Infilling} and \textit{Sentence Shuffling}. The vocabulary size of this PLM ($|V_{\text{PLM}}|$) is $50265$. %

    \item  \textbf{PEGASUS}~\cite{pmlr-v119-zhang20ae}: PEGASUS masks multiple whole sentences that are principle to the document and only generate the masked sentences as a single output sequence. We use the PEGASUS-LARGE\footnote{\url{https://huggingface.co/google/pegasus-large}} model available from the \textit{huggingface} library. PEGASUS has $568$ Million parameters, uses \textit{Sentencepiece-Unigram} tokenization, and its pretraining objective is \textit{Gap Sentence Generation}. The vocabulary size of this PLM ($|V_{\text{PLM}}|$) is $96103$. %
\end{itemize}

\subsection{Datasets}\label{appendix:data_cleaning}
We describe here three details on the target task dataset mentioned briefly in Section~\ref{sec:target-task-dataset}: (i) a detailed description of target task datasets, (ii) training-validation-test data splits, and (iii)the training data cleaning procedure.
\subsubsection{Target Task Dataset Details}
We use four target task datasets in this study: two query-focussed summarization dataset: EBM and BioASQ and two recent benchmark medical question summarization datasets: MeQSum and CHQSum each of which we describe below.
\begin{itemize}
    \item \textbf{EBM}~\cite{DBLP:conf/acl-alta/MollaS11a}. Here input to the system is a query along with a PubMed abstract, and the expected output is the summary answering the question with the PubMed Abstract as the context. %

    \item  \textbf{BioASQ}~\cite{cite:BioASQ}. We use the dataset from BioASQ-9B Phase-B summarization task. The input to the system is a question followed by relevant snippets from a collection of PubMed Abstracts. There are two kinds of outputs an exact answer and an ideal answer associated with the input. For the summarization task, we consider the ideal answer as the Reference summary. %

    \item \textbf{MeQSum}~\cite{ben-abacha-demner-fushman-2019-summarization}. The dataset is created for better medical question summarization because the original patients’ questions are verbose. The dataset contains $1000$ patients’ health questions selected from a collection distributed by the U.S. National Library of Medicine. Each question is annotated with a summarized question by medical experts.

 \item \textbf{CHQSum}~\cite{yadav2022chq}. CHQSum consists of $1507$ domain-expert annotated question-summary pairs from the Yahoo community question answering forum\footnote{\url{https://webscope.sandbox.yahoo.com/catalog.php?datatype=l&did=11}} which provides community question answering threads containing users’ questions on multiple diverse topics and the answers submitted by other users. The authors with the help of $6$ domain experts identified valid medical question from the forum and asked the experts to formulate an abstractive summary for the questions. 
 
\end{itemize}
 
\noindent In case of EBM and BioASQ, for each data point we append the Query and the PubMed Abstracts to form the input SD for the summarization model, and RS is the answer of the datapoint. In case of MeQSum and CHQSum, for each datapoint SD is the healthcare question and RS is the expert annotated summary of the question.

\subsubsection{Train-Validation-Test splits for Target Task Datasets} 
We used the pre-defined train-valid-test split CHQSum provided in the original studies. BioASQ provided only train-test split thus we further split train set into train and validation set. Since the train-validation-test splits for MeQSum and EBM were not provided, we shuffled the dataset and considered a $70/15/15$ data split. 

\subsubsection{Data Cleaning details} 
We perform two data cleaning steps on the training splits of EBM and BioASQ. We remove those data points for which there is (a) no medical-concept-bearing words overlap between RS and SD, and (b) length of RS is more than length of SD. From the EBM dataset having the original size of train split as $1483$, the first step filters out $35 $data points, and the second step filters out $25$ data points, thus resulting in final training set size of $1423$ data points. From BioASQ having the original size of train split as $2420$, first step filters out $0$ data points and second step filters out $805$ data points, resulting in final train split size of $1525$.

\subsection{Evaluation Metrics} \label{appendix:ROUGE}
We first describe the implementation details for computing Rouge scores discussed in Section~\ref{sec:eval-metrics}, where we use the official Rouge~\cite{lin-2004-rouge} script\footnote{\url{https://github.com/bheinzerling/pyrouge/tree/master}}. The following parameters: \textit{-c 95 -2 -1 -U -r 1000 -n 4 -w 1.2 -a}, are used and we report the median at a $95$\% confidence interval. %

However, Rouge fails to match words differing in surface form but belonging to the same medical concept. We thus develop \textbf{MedRouge} (MR) evaluation metric that adds a medical concept normalization step using QuickUMLS~\cite{soldaini2016quickumls} over Rouge; MR-1 and MR-2 are the two variants corresponding to Rouge-1 (R-1) and Rouge-2 (R-2). First, we mark unigrams (for Rouge-1) and bigrams (for Rouge-2) in the generated summaries that have surface form overlap with the reference summaries.  Next, from the remaining set of unigrams and bigrams, we identify those with no surface form overlap but overlap at concept level (e.g., \textit{treatment} and \textit{therapy}) and flag them to consider for the final Rouge-1 and Rouge-2 computation. 

\subsection{Baseline Models} \label{appendix:IFT_Setups_app}
Here, we provide implementation details regarding the vocabulary adaptation and intermediate fine-tuning baseline models as described in Section~\ref{sec:baseline-models}. %

\begin{itemize}
    \item \textbf{IFT-CNN}: In this setup, we take the PLMs models and do the intermediate fine-tuning using CNN/DailyMail news summarization dataset. This acts as a strong baseline against \drift.

    \item \textbf{IFT-PAC}: In this setup, we use the PubMed Abstract Collections (PAC) dataset to perform the task of intermediate fine-tuning. This model is \drift{} without vocabulary adaptation.

    \item \textbf{BSA-BioBERT}: BioBERT~\cite{cite:BioBert} continuously pre-train \textit{bert-base-cased} model on a huge biomedical corpora containing PubMed abstracts, PMC-Articles. We replace the encoder component of the BertSumAbs model which was previously \textit{bert-base-uncased}, with the BioBERT model in this case.

    \item \textbf{BSA-PubMedBERT}: Unlike BioBERT, PubMedBert~\cite{pubmedbert} learns the model vocabulary from scratch and is vastly different from the original Bert vocabulary. Similar to BSA-BioBERT, we replace the encoder component of the BertSumAbs model, with the PubMedBERT model.

     \item \textbf{BSA-AVoCado}: AVocaDo~\cite{hong-etal-2021-avocado} is a work primarily in the classification field. For fair comparison, we only incorporate the vocabulary adaptation module where the subwords to be added to the $V_\text{PLM}$ vocabulary is identified. We then use this updated vocabulary and perform IFT-PAC using PAC as we do in the case of \drift{} for BertSumAbs.
\end{itemize}

\subsection{Hyperparameters}\label{appendix:training-details-app}
We discuss the following hyperparameters in line with discussion in Section~\ref{sec:pac-construct} as follows: (i) the optimal hyperparameters obtained for each target task dataset using \drift{}, (ii) the training hyperparameters, (iii) inference hyperameters.

\subsubsection{Hyperparameter Search for Vocabulary Construction} 
During the vocabulary construction for \drift{} as described in Algorithm~\ref{algo:drift_creator}, we identified two hyperparameters $K$ and $A$ that we obtain for each of the models for each of the target task datasets. We report the optimal values thus obtained in Table~\ref{tab:hyperparams_algo}. We observe that these values vary drastically across different model types.

\begin{table}[h]
    \scriptsize
    \centering
    \setlength{\tabcolsep}{0.05cm}
    \scalebox{0.9}{
    \begin{tabular}{ccccc|cccc|cccc} \hline
                & \multicolumn{4}{c|}{\textbf{BSA}} & \multicolumn{4}{c|}{\textbf{BART}}  & \multicolumn{4}{c}{\textbf{PEGASUS}} \\
           \textbf{Dataset}     &  \multicolumn{4}{c|}{$V_{PLM}$: 30522} &\multicolumn{4}{c|}{$V_{PLM}$: 50265} &\multicolumn{4}{c}{$V_{PLM}$: 96103} \\ 
            & $K$ & $A$ & $|V_{\text{\drift{}}}|$ & Time & $K$ & $A$ & $|V_{\text{\drift{}}}|$ &  Time&$K$ & $A$ & $|V_{\text{\drift{}}}|$ & Time\\ \hline
    EBM & 15K & 0.25 &  32121 & 2 &  15K & 8  &  61326 & 8 & 20K & 10 & 97621 & 14   \\ 
    BioASQ  & 15K & 0.25 &  32941 & 3 &  15K & 5  &  56727 & 8  &20K & 10 & 98721 & 14 \\ 
    MeQSum  & 15K & 0.25 &  30689 & 1 & 10K & 8  &  51012  & 8  & 15K & 1 & 96133 & 9\\ 
    CHQSum & 15K & 0.2 &  30695 & 1 & 10K & 7  &  50945 & 8   & 20K & 1 & 96117 & 7  \\ \hline
    \end{tabular}}
    \caption{The optimal hyperparameter values of Algorithm~\ref{algo:drift_creator}. We then provide the resultant \drift{} vocabulary size and the time required for hyperparameter search for each target dataset and PLM model type in hours.}
    \label{tab:hyperparams_algo}
\end{table}

\subsubsection{Training Hyperparameters} 
All the experiments are run on three $32$ GB Tesla V100 GPUs. We use the training scripts provided by BertSumAbs in their codebase\footnote{\url{https://github.com/nlpyang/PreSumm}}. We use the standard fine-tuning summarization scripts for BART and PEGASUS provided in huggingface codebase\footnote{\url{https://github.com/huggingface/transformers/blob/main/examples/pytorch/summarization/run_summarization.py}}. The final training times for IFT-PAC for \drift{} is mentioned in Table~\ref{tab:runtime-IFTS}. We describe two types of hyperparameters.

\begin{itemize}
\item {\bf Common Hyperparameters: } \textit{number of epochs} for fine-tuning: $5$, \textit{check-pointing}: $2500$ steps, and \textit{accumulation steps}: $10$ (for BertSumAbs) and $2$ (for BART and PEGASUS). 

\item {\bf PLM-specific Hyperparameters: }In case of BertSumAbs there are four additional hyperparameters: learning rate for encoder: $0.002$, learning rate for decoder: $0.01$, warm-up steps for encoder: $20000$, warm-up steps for decoder: $15000$. In case of BART and PEGASUS, we use the huggingface scripts and its associated default hyperparameters, like learning rate: 5e-5 and modify source and target length based on target task datasets for appropriate input truncation to do the fine-tuning. 
\end{itemize}

\begin{table}[h]
    \scriptsize
    \centering
    \begin{tabular}{cccc} \hline
       \textbf{Dataset}  & \textbf{BSA}  & \textbf{BART} & \textbf{PEGASUS} \\ \hline
       EBM      & 30     & 30        & 86  \\
       BioASQ   & 36      & 34        & 78 \\
       MeQSum   & 30      & 28        & 82 \\
       CHQSum   & 27      & 28        & 81 \\ \hline
    \end{tabular}
    \caption{Time required in hours for intermediate fine-tuning  using PAC for each target task dataset and PLM model setting.}
    \label{tab:runtime-IFTS}
\end{table}

\subsubsection{Inference Hyperparameters}
We used beam search to run the \textbf{inference} on the test set. We tuned the following hyperparameters of beam search: beam size (B $\in [2,10]$) and length-penalty~\cite{wu2016googles} ($lp \in (0.1,3]$) on the validation split of the target task dataset. The best values of hyperparameters thus obtained are mentioned in Table~\ref{tab:generation_hyperparameters}.
 
\begin{table}[!ht]
    \scriptsize
    \centering
    \begin{tabular}{ccc|cc|cc} \hline
         \textbf{Dataset} &  \multicolumn{2}{c|}{\textbf{BSA}}&  \multicolumn{2}{c|}{\textbf{BART}}& \multicolumn{2}{c}{\textbf{PEGASUS}}\\ 
           & $B$ & $lp$ & $B$ & $lp$ & $B$ & $lp$ \\ \hline
         EBM &  7 & 2.5 & 2 & 0.7 & 9 & 2.5 \\
         BioASQ  &  7 & 2.5 & 8 & 3 & 9 & 3.5 \\
         MeQSum  &  8 & 0.7 & 6 & 0.3 & 8 & 0.9 \\
         CHQSum &  6 & 0.6 & 6 & 0.3 & 8 & 0.9 \\ \hline
    \end{tabular}
    \caption{Optimal values for inference hyperparameters - beam size ($B$) and Length Penalty ($lp$) used for beam-search generation for each of the PLM against each of the datasets.}
    \label{tab:generation_hyperparameters}
\end{table}

\section{Experimental Results}\label{appendix:results}
Here we describe two things: (i) additional evaluation of \drift{} in extenstion to discussion in Section~\ref{sec:disc-results}, and (ii) human evaluation setup as discussed in Section~\ref{sec:human-evaluation}.
\subsection{Additional Evaluation of \drift}\label{appendix:eval-medvoc}
Here we provide a discussion of \drift{} along two dimensions: (i) how does \drift{} perform against standard vocabulary adaptation baselines like AVocaDo in terms of fragment score,  (ii) performance comparison using MedRouge.
\subsubsection{\drift{} results in better fragment score than AVocaDo}\label{sec:vocab-adapt-app}
We compare the variation in fragment scores observed in \drift{} and BSA-AVocaDo. In MeQSum and CHQSum, the resultant vocabulary size from AVocaDo is more than that of \drift{}, and for EBM and BioASQ the trend is reversed. To understand the effect of vocabulary adaptation and for a fair comparison across differing vocabulary sizes, we first remove the common part from both the vocabularies and make vocabularies of the same size either by randomly sampling or selecting top elements from the larger vocabulary.  We find that in all the cases \drift{} results in a lower (or comparable) fragment score to that of AVocaDo. We report the original fragment scores and the best fragment scores obtained from AVocaDo and \drift{} using the strategy discussed above in Table~\ref{tab:compare_frag_AVocaDo}. This also ascertains our hypothesis of the correlation between optimizing fragment score on downstream target task dataset and target task performance.

\begin{table}[!ht]
    \scriptsize
    \centering
    \setlength{\tabcolsep}{0.15cm}
    \begin{tabular}{ccccc} \hline
     \textbf{Dataset}& \multicolumn{2}{c}{\textbf{Original}} & \multicolumn{2}{c}{\textbf{Overlap Removed}} \\
             & AVocaDo & \drift{} & AVocaDo & \drift{} \\ \hline
     EBM     & 1.53 & 1.51 & 1.55 & 1.53 \\
     BioASQ  & 1.62 & 1.61 & 1.64 & 1.62  \\ 
     MeQSum  & 1.40 & 1.38 & 1.41 & 1.41   \\
     CHQSum  & 1.28 & 1.27 & 1.30 & 1.29   \\
     \hline
    \end{tabular}
    \caption{Fragment score values observed for AVocaDo and \drift{} for BertSumAbs model on three datasets. \textbf{Original} block refers to the fragment score obtained from the original vocabulary of AVocaDo and \drift{}. In \textbf{Overlap Removed} block, we remove the overlapping vocabulary and use an equal-sized vocabulary. We see that in $7$ settings \drift{} results in a $1.07\%$ lower fragment score on average than AVocaDo.}
    \label{tab:compare_frag_AVocaDo}
\end{table}

\subsubsection{\drift{} outperforms baselines in terms of MedRouge}
We observe that the same performance improvement trend as seen with Rouge-L also holds for MedRouge (Table ~\ref{tab:results_pipeline-MedRouge}). There is a performance increase in terms of MedRouge (MR-1 and MR-2) metric of $6.99$\%, $10.88$\%, $7.91$\%, and $1.36$\%  performance improvement over baselines across EBM, BioASQ, MeQSum, and CHQSum datasets respectively. This indicates that even if there is a surface-level mismatch, a major portion of the summaries are meaningful and match at the concept level.

\begin{table}[!ht]
    \scriptsize
    \centering
    \setlength{\tabcolsep}{0cm}
      \begin{tabular}{ccc|cc|cc} \hline
     \textbf{Model}& \multicolumn{2}{c|}{\textbf{BSA}}& \multicolumn{2}{c|}{\textbf{BART}}& \multicolumn{2}{c}{\textbf{PEGASUS}}\\
     & \textbf{MR-1}&\textbf{MR-2}& \textbf{MR-1}&\textbf{MR-2} & \textbf{MR-1}&\textbf{MR-2} \\ \hline
     
     \multicolumn{7}{c}{\textbf{EBMSumm}} \\ 
     
     SOTA &27.00&\secondbest{10.17} & 27.02&\best{13.21} & 26.96&9.96 \\

     IFT-CNN & 28.37&10.04 & 29.50&11.12 & 26.96&9.96 \\
     
     IFT-PAC & \secondbest{29.73}&10.14 & \secondbest{29.80}&11.16 & \secondbest{28.71}&\secondbest{11.26} \\
     
     \drift{}  &  \best{30.09}&\best{11.50}  & \best{31.07}&\secondbest{11.23} &  \best{31.00}&\best{11.86} \\  %

    \hline
    \multicolumn{7}{c}{{\bf BioASQ}} \\ 
    
    SOTA & 50.71&40.55 &  \secondbest{53.15}&\best{43.02} &  46.72&33.25\\

     IFT-CNN & 46.30&36.00 & 49.42&40.30& \secondbest{46.72}&33.25 \\
     
     IFT-PAC & \secondbest{51.31}&\secondbest{41.24} & 51.69&41.09 & 46.68&\best{37.67} \\
     
      \drift{}  & \best{53.75}&\best{43.55}  & \best{{53.94}}&\secondbest{42.56}  & \best{48.79}&\secondbest{36.25}\\  %
     
    \hline
    \multicolumn{7}{c}{{\bf MeQSum}} \\ 
    
     SOTA & \secondbest{57.28}&\secondbest{44.82} & 59.94&46.93 &  58.25&45.33 \\

     IFT-CNN & 50.71&35.38&  \best{63.82}&\best{50.18} & 58.25&45.33\\
    
     IFT-PAC & 54.50&40.58 & \secondbest{63.44}&\secondbest{49.60} & \best{63.12}&\best{50.91} \\
     
     \drift{}  & \best{59.01}&\best{44.88} &  62.34&49.26 & \secondbest{60.41}&\secondbest{47.45}  \\  %
    \hline
    \multicolumn{7}{c}{{\bf CHQSum}} \\
    SOTA & 41.67&22.93 & 46.36&28.44 & 47.86&30.34\\
    
    IFT-CNN &  42.54&24.20 &  \secondbest{47.05}&\best{29.53} & 47.86&\secondbest{30.34} \\
    
    IFT-PAC  & \secondbest{43.78}&\best{25.03} & 46.94& \secondbest{29.52} &  \secondbest{48.04}&30.24\\ 
    
    \drift{}  &  \best{44.14}&\secondbest{24.90} &  \best{47.77}&29.30 &  \best{48.10}&\best{30.41} \\  \hline
    \end{tabular}
    \caption{Median MedRouge values for \drift{}, baselines and SOTA models. We find that \drift{} outperforms baselines and even SOTA in majority of the cases.} %
    \label{tab:results_pipeline-MedRouge}
\end{table}

\subsection{Human Evaluation}\label{appendix:human-eval-app}
The annotations were conducted on the Prolific\footnote{\url{https://www.prolific.com/}} platform using $30$ participants in total. Each participant was shown $10$ random samples from a pool of $100$ summary pairs (order of summary randomized and anonymized) and was given $30$ minutes to complete the study. We also collected demographic information and annotation experience feedback from the participants. The study instrument was designed using the Potato tool~\footnote{\url{https://github.com/davidjurgens/potato}}. The participants were compensated at the rate of $9$ pounds/hour, which according to platform guidelines was fair compensation. Proper consent notice was shown to the participants and no personal information (other than age) was collected in the demographic information. The filtering criteria for participants were kept as follows: 
\begin{itemize}
    \item \textbf{Age:} $\ge 25$,
    \item \textbf{Primary Language:} English,
    \item \textbf{Highest education level completed:} Graduate degree (MA/MSc/MPhil/other), Doctorate degree (PhD/other)
    \item \textbf{Subject:} Medicine, Health and Medicine, Biomedical Sciences.
\end{itemize}
The annotations were carried across three dimensions~\cite{fabbri-etal-2021-summeval} each of which we discuss below. 
\begin{itemize}
    \item \textbf{Coherence}. The summary should be well-structured and well-organized. The summary should not just be a heap of related information but should build from sentence to sentence to a coherent body of information. 

    \item  \textbf{Relevance}. The summary should include only important information from the source document. In the case of a query, you must also judge how relevant is the summary to the query based on the given source document.

    \item \textbf{Faithfulness}. A faithful summary contains only statements that are entailed by the source document. You may also penalize summaries that contain facts not supported (or can not be verified) in the source document (termed as hallucinated facts). 

\end{itemize}

For each of these dimensions, we show one positive (high rating) and one negative example (low rating) along with an explanation  as a part of our annotation guideline (Table~\ref{tab:sample-annot-guideline}).

\begin{table}
    \scriptsize
    \centering
    \begin{tabular}{p{1.25cm}p{6.5cm}}
    \hline
         \textbf{Source Document}& KNEE OSTEOARTHRITIS. \\
            & Good morning about 20 years ago I suffered ruptured anterior cruciate ligament and removal of domestic law meniscus, was operated and made me clancy, at present unfortunately my knee is totally affected and I have arthritis and severe pain, according to a dr traumatologo commented me I need a knee prosthesis my question is can you treat me, or turn me can recommend doctors or hospitals to treat in the U.S. \\ \hline
         \multicolumn{2}{c}{\textbf{Positive Example}}\\ \hline
         \textbf{Summary} & How can I find physician(s) or hospital(s) who specialize in knee osteoarthritis?\\
         \textbf{Rating} & $5$\\
         \textbf{Explanation} & Here we can see the summary is focused on knee osteoarthritis and asks how to find physicians or hospitals who specialize in it. \\ \hline
         \multicolumn{2}{c}{\textbf{Negative Example}}\\ \hline
         \textbf{Summary} & What are the treatments for ruptured anterior cruciate ligament and meniscus?\\
         \textbf{Rating} & $2$ \\
         \textbf{Explanation} & In the source document the patient is asking for recommendations for hospitals or doctors who specialize in the treatment for the topic of knee osteoarthritis.\\ \hline
         
    \end{tabular}
    \caption{A negative and positive example as shown to the participant in the annotation guidelines for clarification under \textit{Relevance} dimension of annotation. The data point is taken from MeQSum dataset.}
    \label{tab:sample-annot-guideline}
\end{table}

\noindent \textbf{Demographic analysis of participants.} The average age of participants was $30$ years. Out of $30$ participants, $70$\% were female and  $30$\% were male. $90$\% were Graduates, and $10$\% were PhD holders. $60$\% of the participants were practicing medicine, or previously practiced in a clinical setting. The participants came from $11$ different countries, majority of which came from UK ($n=11$) and US ($n=4$).

\noindent \textbf{Annotation experience of participants.} After the study, we took feedback from the participants regarding the overall annotation experience. Our primary focus was on three aspects: (i) Clarification of the annotation guidelines, (ii) Usability of Potato, (iii) Overall experience with the study setup. $60$\% of participants found the annotation guidelines to be mostly clear, and $30$\% found it very clear. $80$\% of the participants were satisfied with the Potato platform interface.  $90$\% of the participants were overall satisfied with the study design and summary pairs shown to them.

\section{Limitations}
First, the evaluation of \drift{} in this work is limited only to encoder-decoder-based PLMs, we thus plan to extend and evaluate \drift{} on decoder-only-based PLMs like GPT. Second, in \drift{} we identify medical-concept-bearing using the QuickUMLS tool which uses certain heuristics to identify such words. Although we maintain a very high threshold, there will always be some errors that might cascade through the entire pipeline. Third, we observe that in the case of \drift{} although the speedup achieved is very high, owing to the nature of efficient hyperparameter search. However, the drawback here is that we have to perform one iteration of IFT per target task dataset which is still a huge cost and we will try to alleviate this problem in future iterations. Fourth, although the vocabulary adaptation pipeline of \drift{} is quite flexible to adapt to any domain with high vocabulary mismatch, we do not evaluate the generalizability of \drift{} over non-medical domains such as legal text and scientific literature of rarer subjects. %

\section{Ethics Statement and Broader Impact}\label{sec:ethics}
Summarization systems, in general, any generation systems are prone to hallucination leading to the generation of less faithful summaries. Furthermore, our human evaluation pointed out that \drift{} generates significantly more fraction of faithful summaries when compared to existing baselines. Our view is that summaries produced by such PLMs are not yet production-ready for their intended users like medical professionals, and clinicians. More thorough research is needed to better characterize the kinds of errors (specifically in the context of faithfulness and relevance) made by these PLMs and ultimately to mitigate them.

\end{document}